\documentclass[10pt,twocolumn,letterpaper]{article}

\usepackage[algorithms]{wacv}      % To produce the REVIEW version for the algorithms track
\usepackage{graphicx}
\usepackage{booktabs}
\usepackage{subcaption}
\usepackage{tabularx}
\usepackage{amsmath}
\usepackage{amssymb}
\newcolumntype{C}{>{\centering\arraybackslash}X}
\usepackage{capt-of,etoolbox}

\newcommand{\rev}[1]{\textcolor{black}{#1}}
\usepackage[pagebackref,breaklinks,colorlinks]{hyperref}
\usepackage{orcidlink}
\usepackage{etoolbox}
\usepackage{enumitem}

\usepackage{amssymb}
\usepackage{pifont}
\usepackage[normalem]{ulem}
\newcommand{\cmark}{\ding{51}}%
\newcommand{\xmark}{\ding{55}}%
\usepackage[accsupp]{axessibility} % Improves PDF readability for those with disabilities.

\usepackage[capitalize]{cleveref}
\crefname{section}{Sec.}{Secs.}
\Crefname{section}{Section}{Sections}
\Crefname{table}{Table}{Tables}
\crefname{table}{Tab.}{Tabs.}

\makeatletter
\renewcommand{\paragraph}{%
  \@startsection{paragraph}{4}%
  {\z@}{0.5em}{-0.5em}%
  {\normalfont\normalsize\bfseries}%
}
\makeatother

\def\wacvPaperID{774} % *** Enter the WACV Paper ID here
\def\confName{WACV}
\def\confYear{2025}

\begin{document}

\title{
\begin{minipage}{.05\textwidth}
\centering
\vspace{-4pt}
\includegraphics[width=\linewidth]{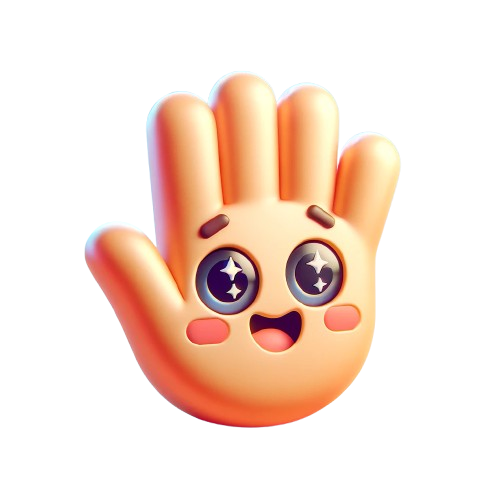} 
\end{minipage}
\hspace{-8pt}
HandCraft: Anatomically Correct Restoration of Malformed Hands \\in Diffusion Generated Images
}

\author{
% \vspace{2mm} 
%\hspace{-4.5mm}
        Zhenyue Qin$^1$\thanks{~~Equal contribution}\hspace{4pt}\thanks{~~Corresponding authors: dylan.campbell@anu.edu.au, \\kf.zy.qin@gmail.com}\hspace{4pt}\thanks{~~Formerly with Seeing Machines }\hspace{3mm}
        Yiqun Zhang$^2$\footnotemark[1] \hspace{3mm}
        Yang Liu$^1$\footnotemark[3] \hspace{3mm}
        Dylan Campbell$^2$\footnotemark[2]
    \\
    \hspace{-3mm}
    \textsuperscript{1}Seeing Machines  
    \hspace{4mm}  
    \textsuperscript{2}Australian National University
    \\
    % {\tt \href{mailto:qingyu.chen@yale.edu}{qingyu.chen@yale.edu}}
    \
Project Page: \url{https://kfzyqin.github.io/handcraft/}
}

\maketitle

%%%%%%%%% ABSTRACT
\begin{abstract}
Generative text-to-image models, such as Stable Diffusion, have demonstrated a remarkable ability to generate diverse, high-quality images. However, they are surprisingly inept when it comes to rendering human hands, which are often anatomically incorrect or reside in the ``uncanny valley''. In this paper, we propose a method HandCraft for restoring such malformed hands. 
This is achieved by automatically constructing masks and depth images for hands as conditioning signals using a parametric model, allowing a diffusion-based image editor to fix the hand's anatomy and adjust its pose while seamlessly integrating the changes into the original image, preserving pose, color, and style.
Our plug-and-play hand restoration solution is compatible with existing \rev{pretrained} diffusion models, and the restoration process facilitates adoption by eschewing any fine-tuning or training requirements \rev{for the diffusion models}. We also contribute MalHand datasets that contain generated images with a wide variety of malformed hands in several styles for \rev{hand detector} training and \rev{hand restoration} benchmarking, and demonstrate through qualitative and quantitative evaluation that HandCraft not only restores anatomical correctness but also maintains the integrity of the overall image.
\end{abstract}

\begin{figure*}[htbp]
\centering
\vspace{-10pt}
\begin{subfigure}{0.21\textwidth}
\centering
\includegraphics[width=\linewidth]{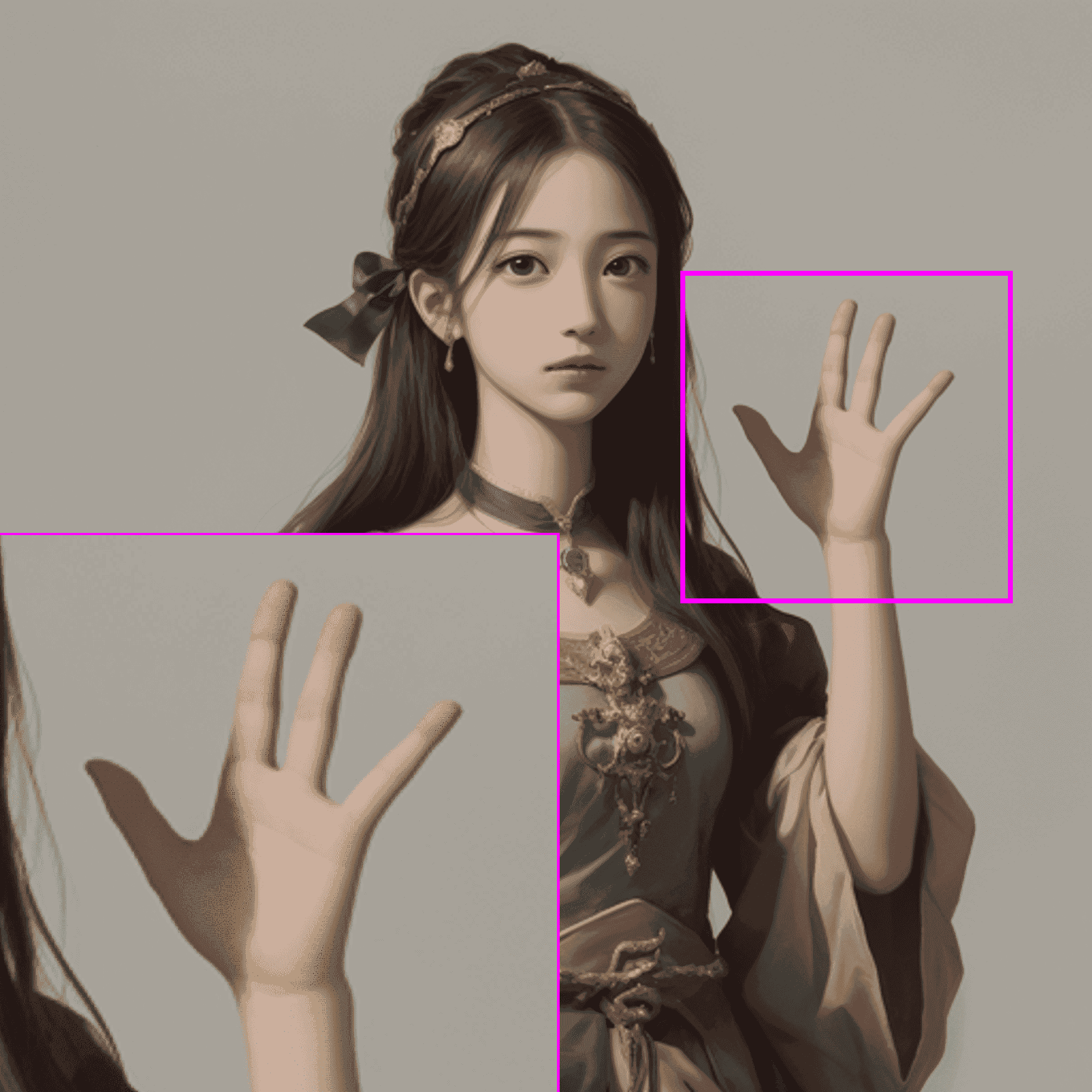}
\label{fig:sub1}
\end{subfigure}%
\hspace{1mm}
\begin{subfigure}{0.21\textwidth}
\includegraphics[width=\linewidth]{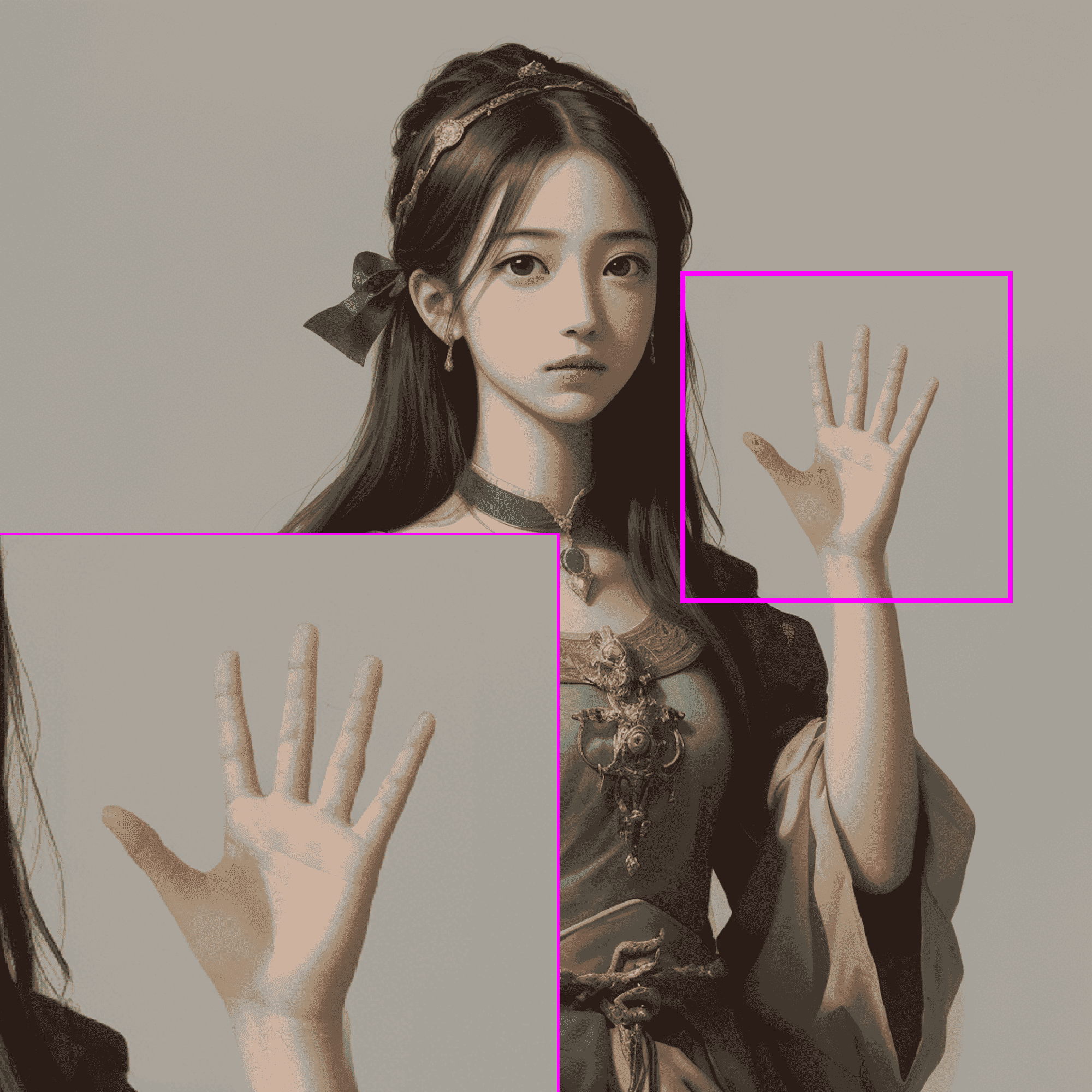}
\label{fig:sub2}
\end{subfigure}%
\hspace{1mm}
\begin{subfigure}{0.21\textwidth}
\centering
\includegraphics[width=\linewidth]{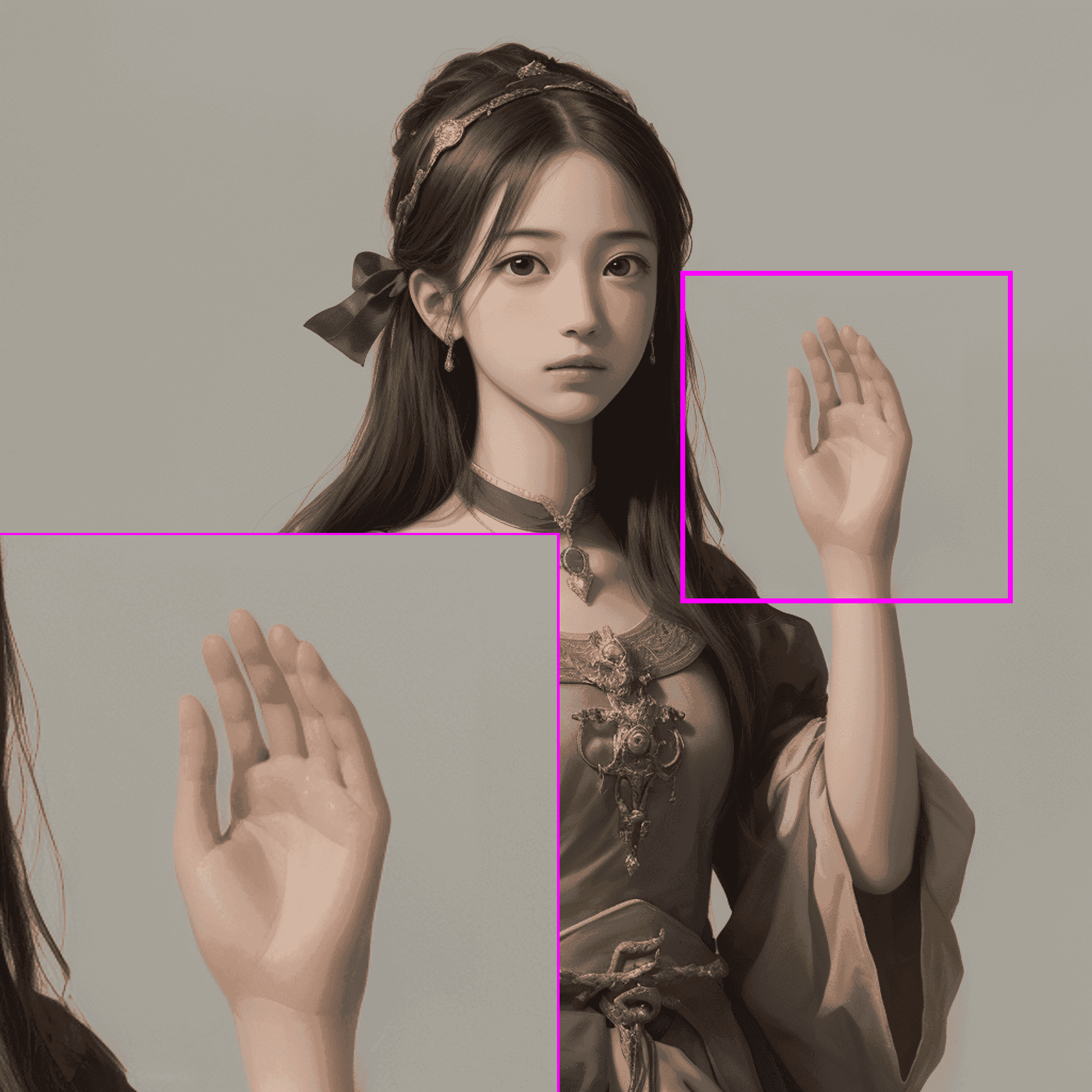}
\label{fig:sub3}
\end{subfigure}%
\hspace{1mm}
\begin{subfigure}{0.21\textwidth}
\centering
\includegraphics[width=\linewidth]{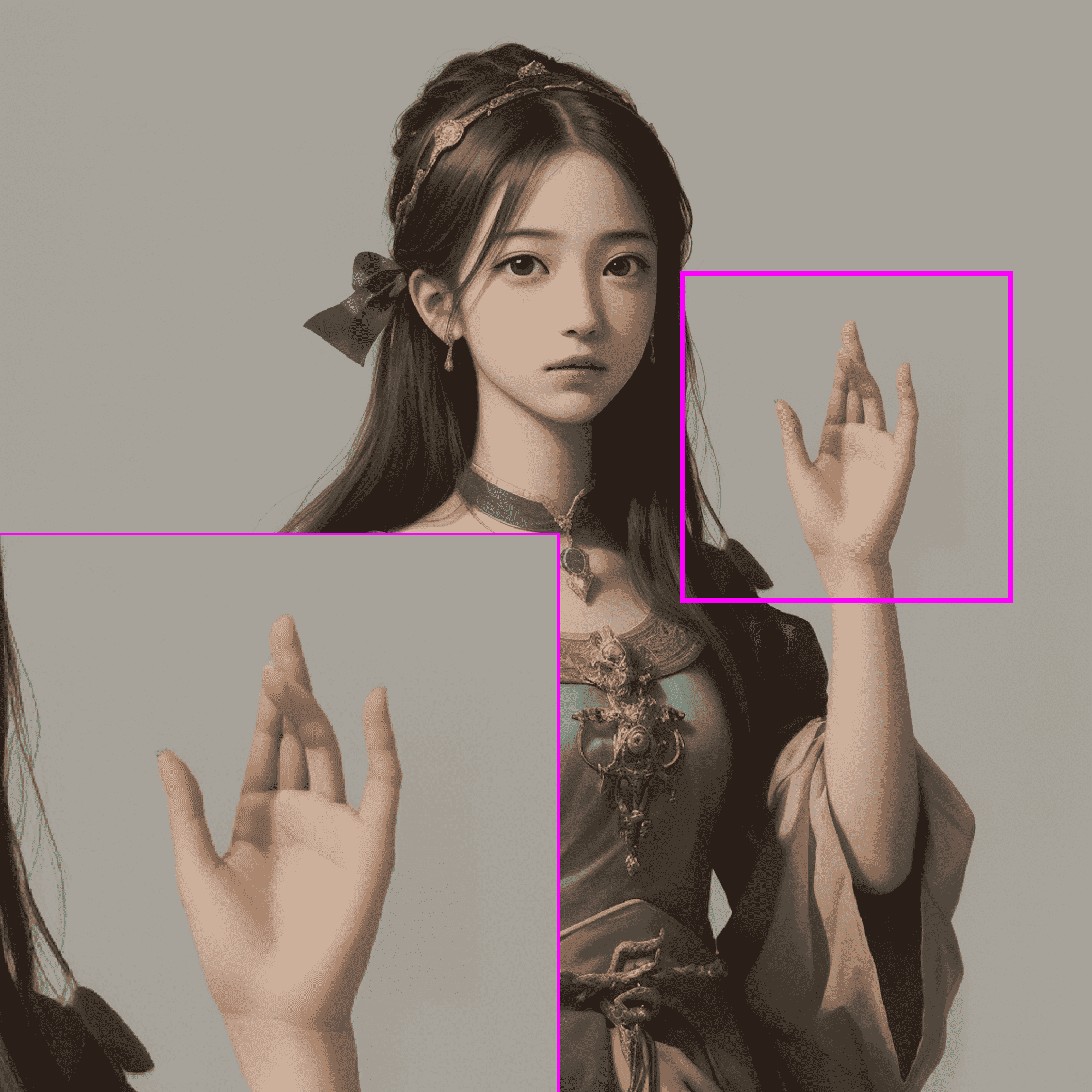}
\label{fig:sub4}
\end{subfigure}%
\vspace{-4mm} % Control the space between the rows here
\begin{subfigure}{0.21\textwidth}
\centering
\includegraphics[width=\linewidth]{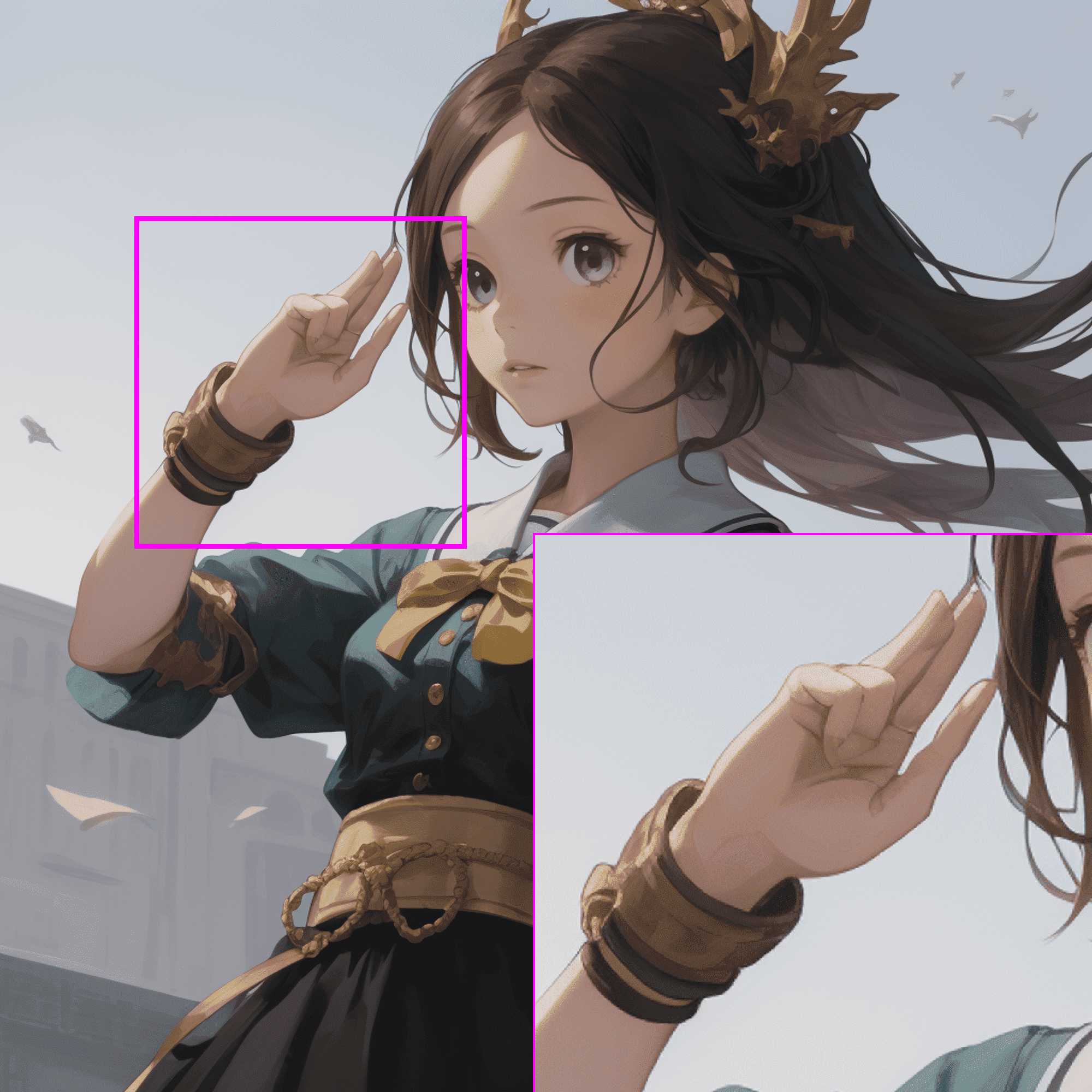}
\caption{Malformed Hand}
\label{fig:sub5}
\end{subfigure}%
\hspace{1mm}
\begin{subfigure}{0.21\textwidth}
\includegraphics[width=\linewidth]{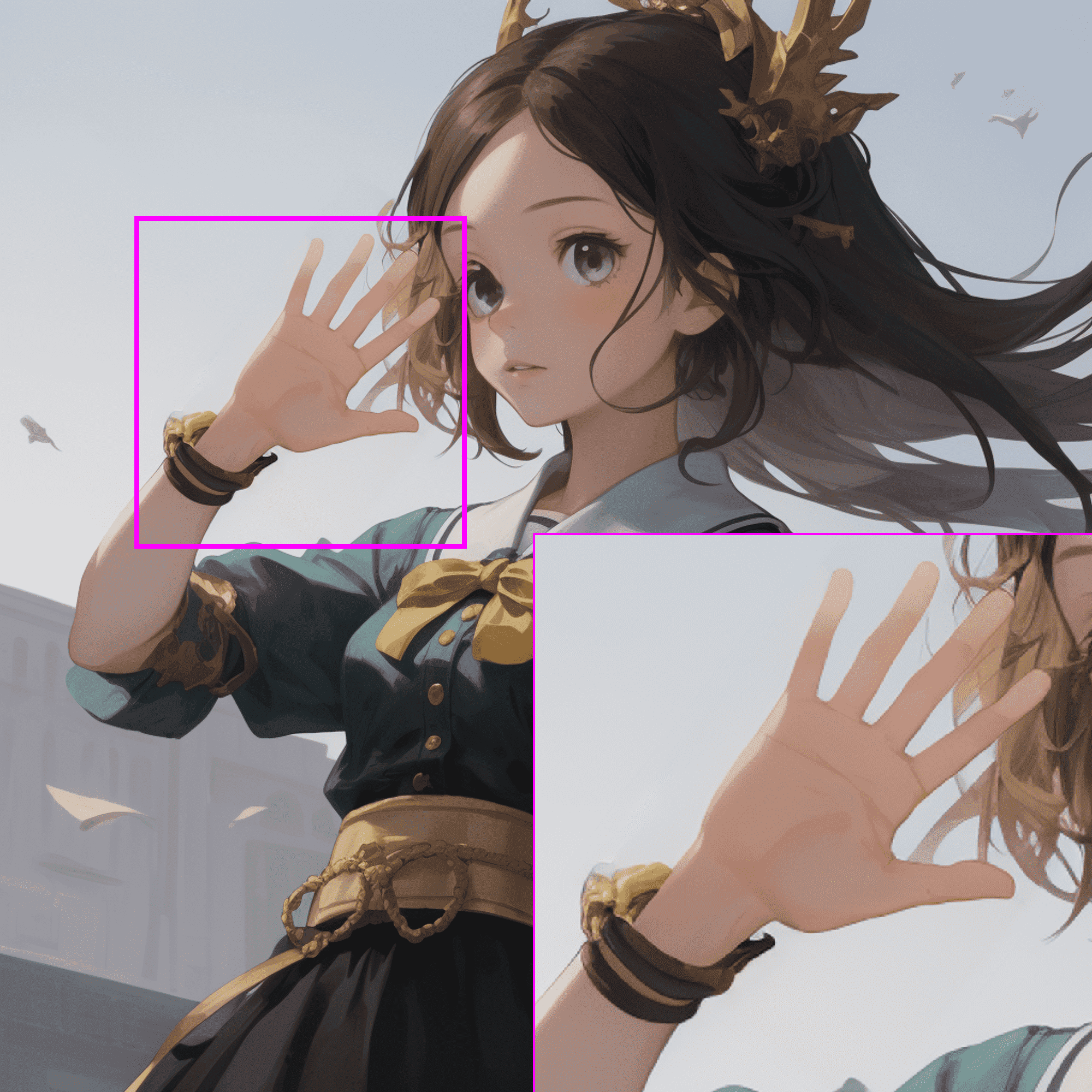}
\caption{Spread Palm}
\label{fig:sub6}
\end{subfigure}%
\hspace{1mm}
\begin{subfigure}{0.21\textwidth}
\centering
\includegraphics[width=\linewidth]{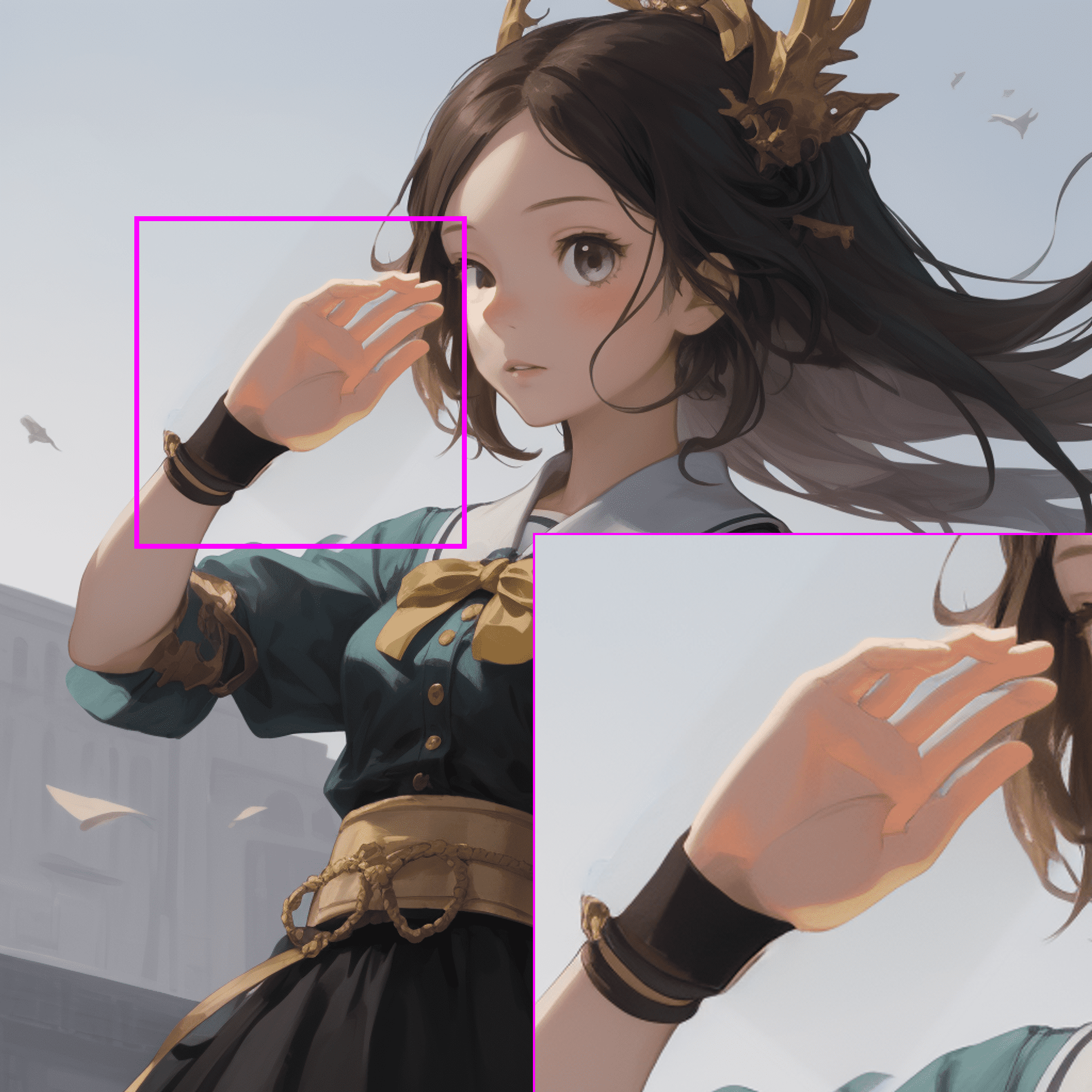}
\caption{Curved Gesture}
\label{fig:sub7}
\end{subfigure}%
\hspace{1mm}
\begin{subfigure}{0.21\textwidth}
\centering
\includegraphics[width=\linewidth]{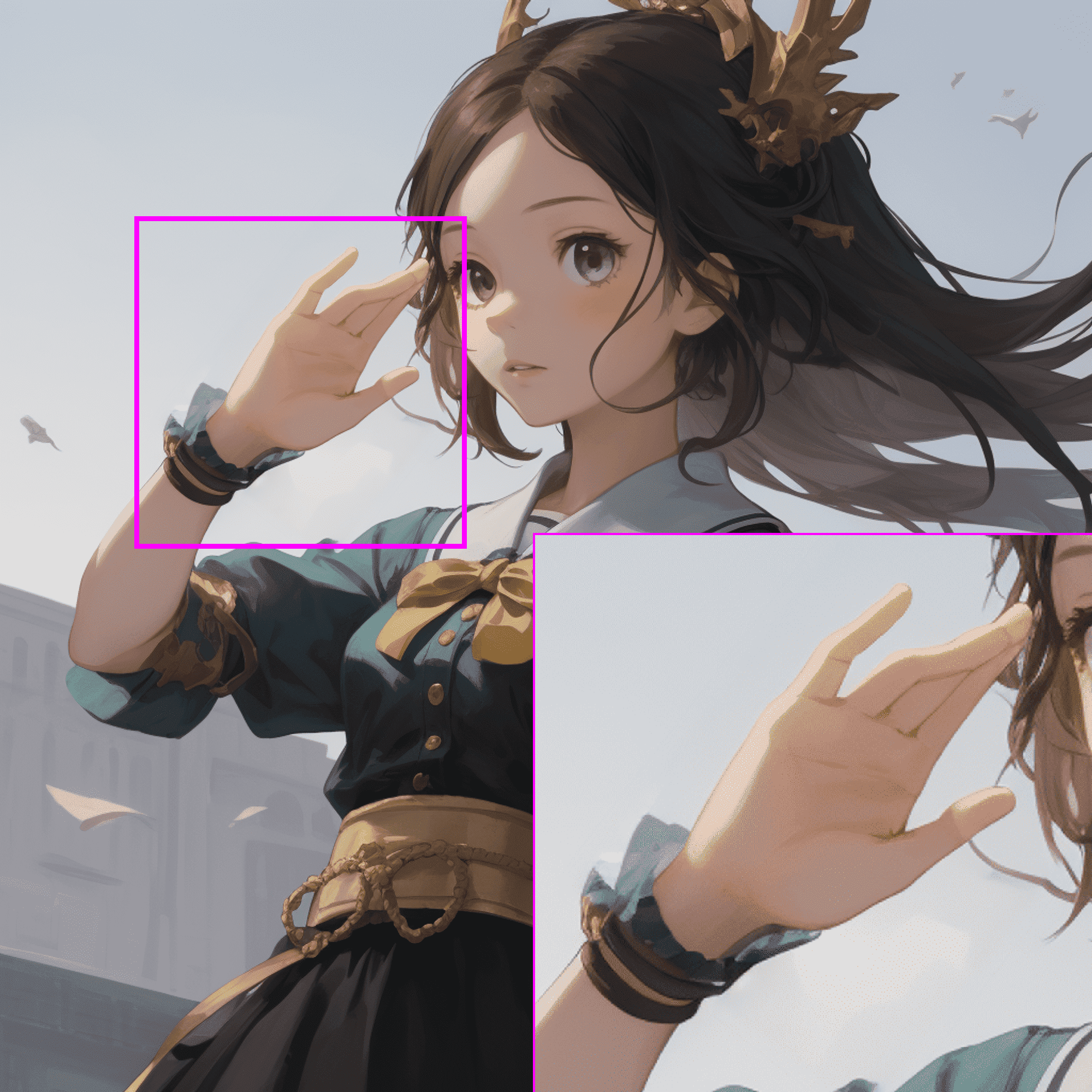}
\caption{Relaxed Pose}
\label{fig:sub8}
\end{subfigure}
\caption{
Images generated by Stable Diffusion~\cite{ldm_2022} often exhibit anatomically incorrect hands (a), for example, a missing finger (top) or abnormal relative finger lengths (bottom). Our method---HandCraft---is able to correct the hands in a controllable manner, allowing for a variety of output gestures while following the style of the original image (b--d). The resulting images feature naturally-posed hands, improving the quality of the AI-generated portraits and restoring the illusion of reality.
}
\label{fig:intro_img}
\end{figure*}

%%%%%%%%% BODY TEXT

\section{Introduction}
\label{sec:intro}

Text-to-image diffusion models, such as Stable Diffusion \cite{ldm_2022}, have gained wide popularity due to their remarkable capability to generate diverse, high-quality images across a wide range of styles \cite{diffusion_survey_2023,dit_2023}. However, they struggle to accurately render human hands, often producing anatomically incorrect or highly unusual forms~\cite{handdiffuser_2024}. These errors can include hands with supernumerary or missing digits, atypical relative finger lengths, and other distortions. \cref{fig:intro_img} illustrates two cases of such malformed hands, with a missing finger in the top row and abnormal relative finger lengths in the bottom row. These examples highlight the discrepancy between the generated depictions and human anatomy.

Due to humans' high sensitivity to deviations from the expected human form, generating malformed hands often leads to an ``uncanny valley''~\cite{uncanny_2012} effect, which affects the realism of these images.
% inhibits any suspension of disbelief. 
This in turn hinders the use of these models as artistic tools.
We note here that we do not use the term ``malformed'' in the pejorative sense, since we recognize that a wide variety of hand shapes are naturally present in the human population or may arise from misadventure. That is, the model is inadvertently forming the hands atypically, rather than intentionally depicting the difference that exists in the human population.

Diffusion models' propensity for generating malformed hands has been widely recognized~\cite{handrefiner_2023, handdiffuser_2024,hand_fix_youtube_2024}. There has been growing interest for techniques to repair these malformed hands, reflected by a large number of tutorials across various languages for this purpose~\cite{openai_fix_hand_2024,hand_fix_civitai_2023,hand_fix_pixcore_2024,sd_hand_youtube_chinese}. However, the restoration methods proposed in these tutorials often necessitate human intervention. For instance, repeatedly inpainting the manually-annotated affected areas until a satisfactory outcome is achieved~\cite{hand_fix_youtube_template_2023}. The requirement for human involvement makes the correction process laborious. Prompt engineering has also emerged as a popular strategy to mitigate the issue of malformed hands in images generated by diffusion models~\cite{chi_prompt_engineering_2022,prompt_engineering_diffusion_2024}. By meticulously designing and refining text prompts, users attempt to guide the model towards generating more anatomically accurate hands~\cite{hand_negative_prompts_github_2024}. Despite these efforts, even well-crafted prompts often fail to prevent the occurrence of malformed hands~\cite{bad_hand_civitai_2024}. 

We introduce an end-to-end framework designed to repair malformed hands in generated images while minimizing the need for human intervention. To achieve this, we propose an approach for generating a hand shape as a conditioning image to guide ControlNet~\cite{controlnet_2023}, a diffusion-based image editing method, in correcting malformed hands. Our method is capable of responsively adjusting the size and angle of the hand shape, ensuring that the restored hand seamlessly integrates with the original human figure, while preserving the surrounding regions of the image unaltered. Experimental results demonstrate the robustness of our approach. Furthermore, our restoration process is designed to be plug-and-play, requiring no further fine-tuning or training \rev{of the ControlNet}, and is therefore easy to integrate into various diffusion models. 
Our contributions are
\begin{enumerate}[noitemsep,nolistsep]
    \item HandCraft, a framework for detecting and restoring malformed hands generated by diffusion models while minimizing alterations to other image regions;
    
    \item a simple yet robust control image generation method to construct a mask and an aligned depth image for the hand region as condition signals, enabling a diffusion-based image editor to restore malformed hands; and

    \item the MalHand datasets, comprising portraits with malformed hands across diverse styles, that can be used to train a malformed hand detector and thoroughly evaluate baseline models.
\end{enumerate}
HandCraft achieves state-of-the-art performance on both the MalHand-realistic and MalHand-artistic datasets.

\section{Related Work}

In this section, we provide a brief overview of image synthesis and editing techniques before discussing approaches for restoring malformed hands in generated images.

\paragraph{Image Synthesis.}
After earlier successes with Variational Autoencoders (VAEs)~\cite{vae_2013} and Generative Adversarial Networks (GANs)~\cite{gan_2014}, diffusion models~\cite{ho2020denoising} have emerged as a powerful new class of generative models.
They are characterized by their ability to map noise into complex images through a gradual denoising process.
This technique was refined by the development of Latent Diffusion Models (LDMs) \cite{ldm_2022}, which tackle the computational challenges by operating in a latent space, significantly improving both efficiency and the quality of generated images.
By leveraging pretrained autoencoders, LDMs offer a versatile and flexible architecture that supports a wide range of conditioning inputs, such as text descriptions. 
This advance enables efficient and adaptable image synthesis models like Stable Diffusion~\cite{ldm_2022}. However, despite the impressive capabilities of these models, generated images of humans often exhibit malformed hands. The complex structure and fine details of hands pose a challenge for these models, often resulting in anatomically incorrect hand representations~\cite{handdiffuser_2024}.

\paragraph{Image Editing}
has become a key application of generative models, allowing users to modify images as desired. Early methods used encoder-decoder architectures to encode, manipulate, and decode images~\cite{globally_2017, generative_inpainting_2018}. Recent techniques employ GANs for editing~\cite{cyclegan_2017, img2img_2018}. ControlNet~\cite{controlnet_2023} uses diffusion models with spatially-localized conditioning to enhance editing, allowing conditions like Canny edges and human poses to guide image generation and editing.

\paragraph{Malformed Hand Restoration.}

Contemporary research addresses correcting malformed hands in Stable Diffusion-generated images. HandDiffuser~\cite{handdiffuser_2024} generates humans with correct hands from text. HandRefiner~\cite{handrefiner_2023} modifies entire images to fix hands, potentially altering unintended aspects. Our research, however, targets only the malformed hand area, aiming for precise corrections with minimal impact on the rest of the image, thus preserving the artwork's integrity and originality.
\begin{figure*}[!t]
    \centering
    \includegraphics[width=0.9\linewidth]{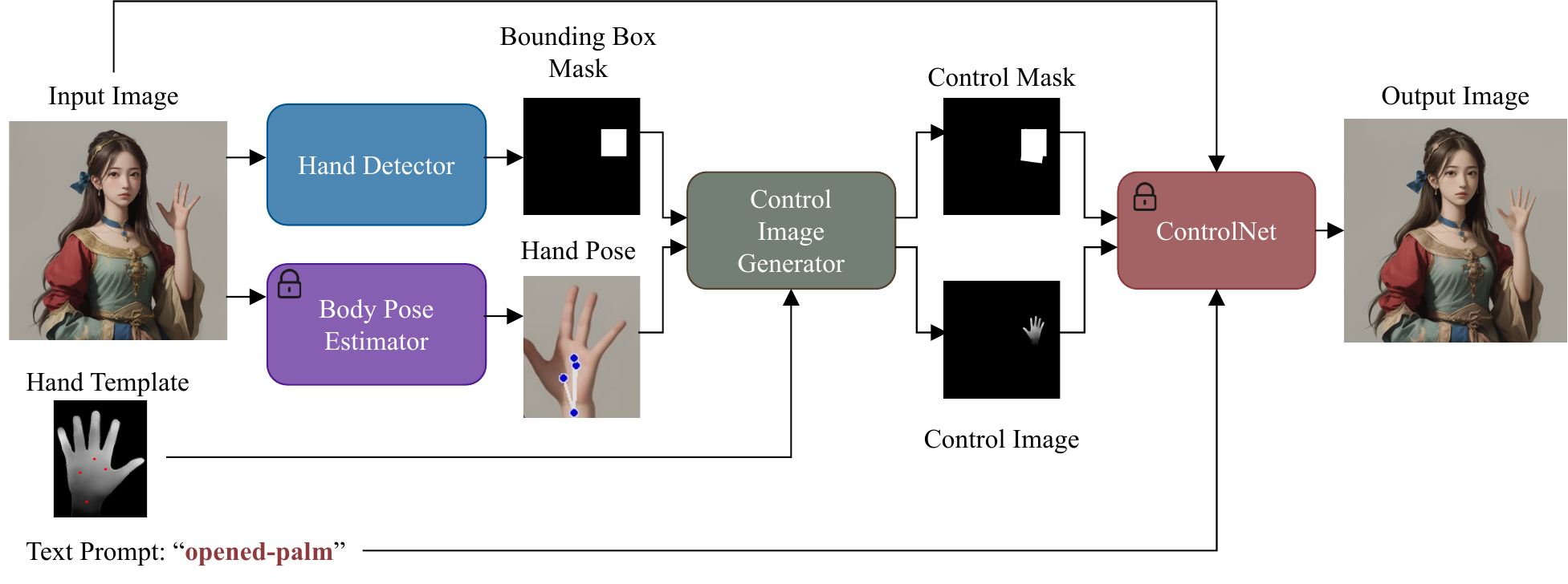}
    \vspace{-6pt}
    \caption{HandCraft flowchart. The framework has three stages for correcting malformed hands in images. (1) Hand detection. A hand detector is employed to detected the bounding box of the hand and a body pose estimator is used to predict the landmarks on hands with the prior of the whole body pose. (2) Control image generation. The extracted body pose and a parametric hand template are given to a control image generator to obtain a control image $I_c$ and a template mask $M_t$. The final control mask $M$ is obtained by doing a union operation between the bounding box mask $M_d$ and the template mask $M_t$. (3) Hand restoration. The final output image with corrected hand is generated using ControlNet given the input image, a text prompt, control mask and control image as the conditioning.}
    \label{fig:flow_chart}
\end{figure*}

\section{Detecting and Restoring Malformed Hands}

In this section, we detail the proposed HandCraft framework. We begin by establishing the notation and providing an overview of the framework's pipeline. Subsequently, we delve into the generation of a conditional hand shape that ensures anatomical plausibility and accurate positioning. This is followed by a discussion on defining the restoration region within the image for localized correction while preserving the overall image integrity. 

\subsection{The HandCraft Framework}
\label{sec:overview_framework}

Our HandCraft framework is designed to address restoring malformed hands in images generated by diffusion models. \rev{Additionally, this framework can also edit well-formed hands in images, allowing us to modify them into desired hand gestures. As illustrated in \cref{fig:flow_chart},} this framework consists of three stages: \rev{hand detection,} 
% control image generation, 
\rev{conditioning signals generation,}
and hand restoration.

At the hand detection stage, a \rev{hand detector} is applied to the input image $I$ to identify the region of interest, producing a bounding box mask $M_d$ for the hand. 
Any pretrained hand detection model, such as YOLOv8~\cite{yolov8}, can be used as the hand detector, but both standard and malformed hands will be detected.
\rev{If we want to avoid modifying the images when the generated hand is not malformed, we can fine-tune the hand detector on stylized images to only detect malformed hands.}
In addition to the hand detector, a body pose estimator (Mediapipe~\cite{mediapipe_2019}) is also used to predict the body pose $S$, which \rev{includes the hand keypoints that} facilitate the correct positioning and orientation of the hand template $T$. 
\rev{Mediapipe exhibits robust capability in estimating keypoints of malformed hands, despite not being specifically designed for this purpose. Its strength lies in leveraging estimations of other body parts to inform hand keypoint predictions, allowing for reasonable estimates even in cases of malformed hands. To assess Mediapipe's success rate in detecting malformed hand keypoints, we conducted an experiment. We randomly selected 500 images containing malformed hands and applied Mediapipe for hand keypoint detection. The results show that Mediapipe successfully provided keypoints for all malformed hands, demonstrating its effectiveness in this challenging scenario.}

At the 
% control image
\rev{conditioning signals}
generation stage, the primary objective is to create a control image $I_c$ \rev{that guides the hand restoration process} and a corresponding control mask $M$ \rev{which defines the restoration region}. To this end, a control image generator aligns a pre-defined hand template $T$, which is a depth map of a hand, using the extracted body pose $S$ to create the control image $I_c$, as well as a template mask $M_t$. The control mask $M$ is obtained by the union of the bounding box mask $M_d$ extracted from the original image and the hand template mask $M_t$, to precisely localize the hand. The depth image $I_c$ and the mask $M$ for hand are crucial conditioning signals for the editing process to achieve a seamless integration of the restored hand.

The final stage is hand restoration. A pretrained ControlNet~\cite{controlnet_2023} model with frozen weights is provided with the control image $I_c$ and mask $M$ to adjust the input image $I$, given text prompt $P$ that describes the shape of the hand template $T$. This restoration process \rev{is training-free and thus compatible with various diffusion models.} It focuses only on the hand region, while preserving the integrity of the rest of the image. The output of this stage is the restored image $I'$, where the malformed hand has been restored to match the desired shape and pose specified by the hand template and text prompt. The restored hand blends with the original image's style and aesthetics, resulting in a more realistic and anatomically correct representation.

Our framework's versatility is evidenced by its successful application to various instances of Stable Diffusion models, demonstrating its efficacy across diverse image styles.

\begin{figure}[!t]
\centering
\vspace{-5pt}
\begin{subfigure}[b]{0.3\linewidth}
    \includegraphics[trim=0 0 0 0, clip, width=\linewidth]
    {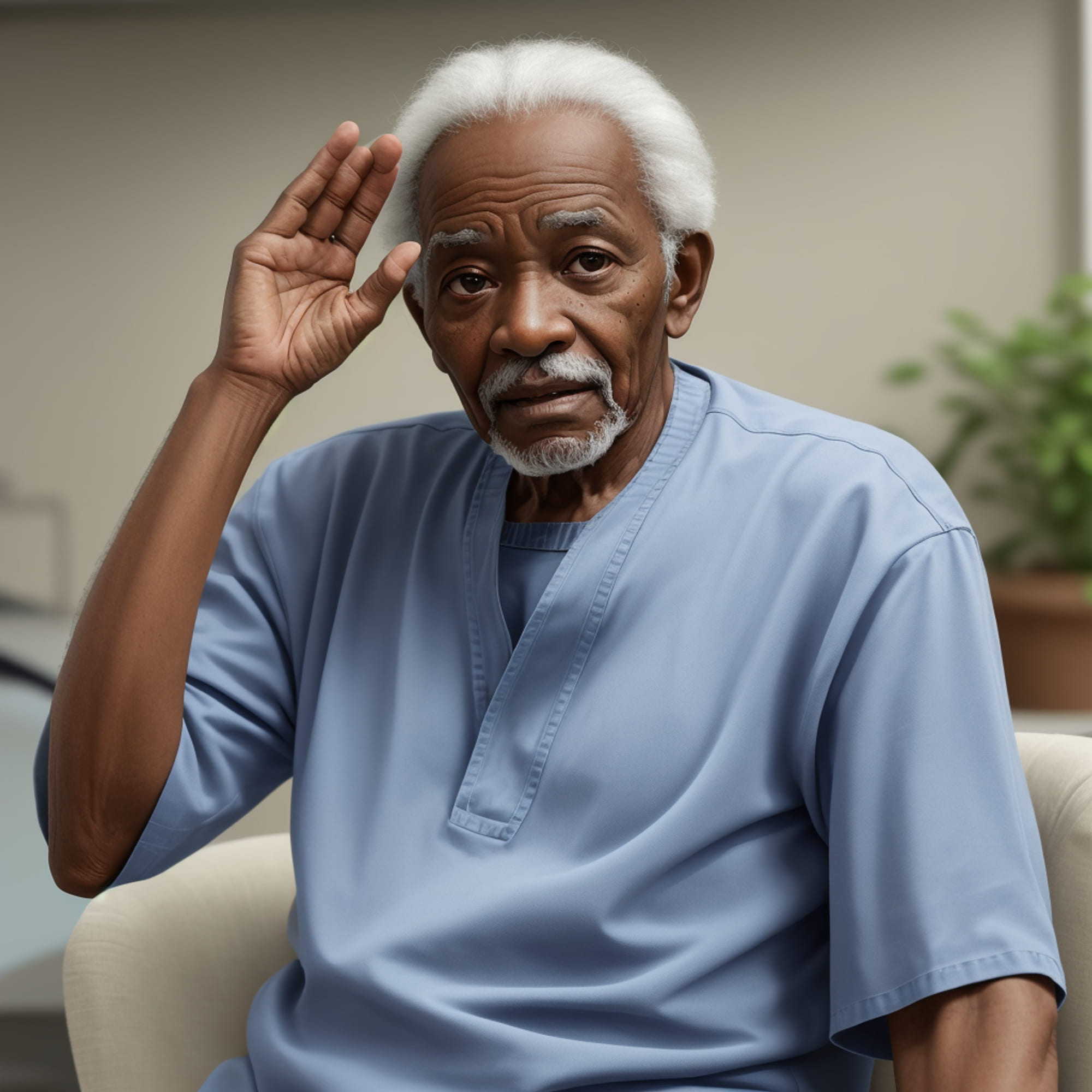}
    \caption*{Input}
\end{subfigure}
\hfill
\begin{subfigure}[b]{0.3\linewidth}
    \includegraphics[trim=0 0 0 0, clip, width=\linewidth]{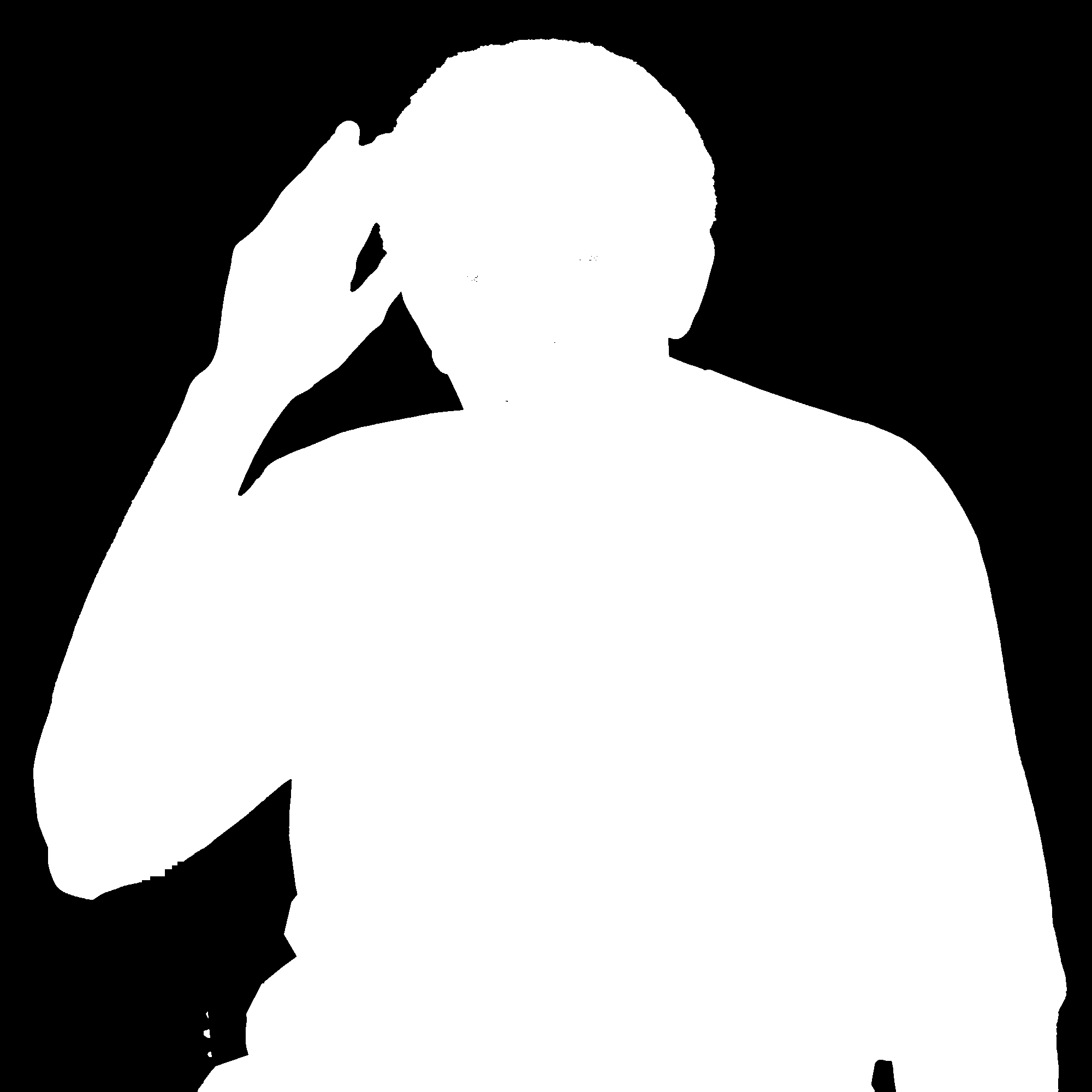}
    \caption*{Silhouette}
\end{subfigure}
\hfill
\begin{subfigure}[b]{0.3\linewidth}
    \includegraphics[trim=0 0 0 0, clip, width=\linewidth]{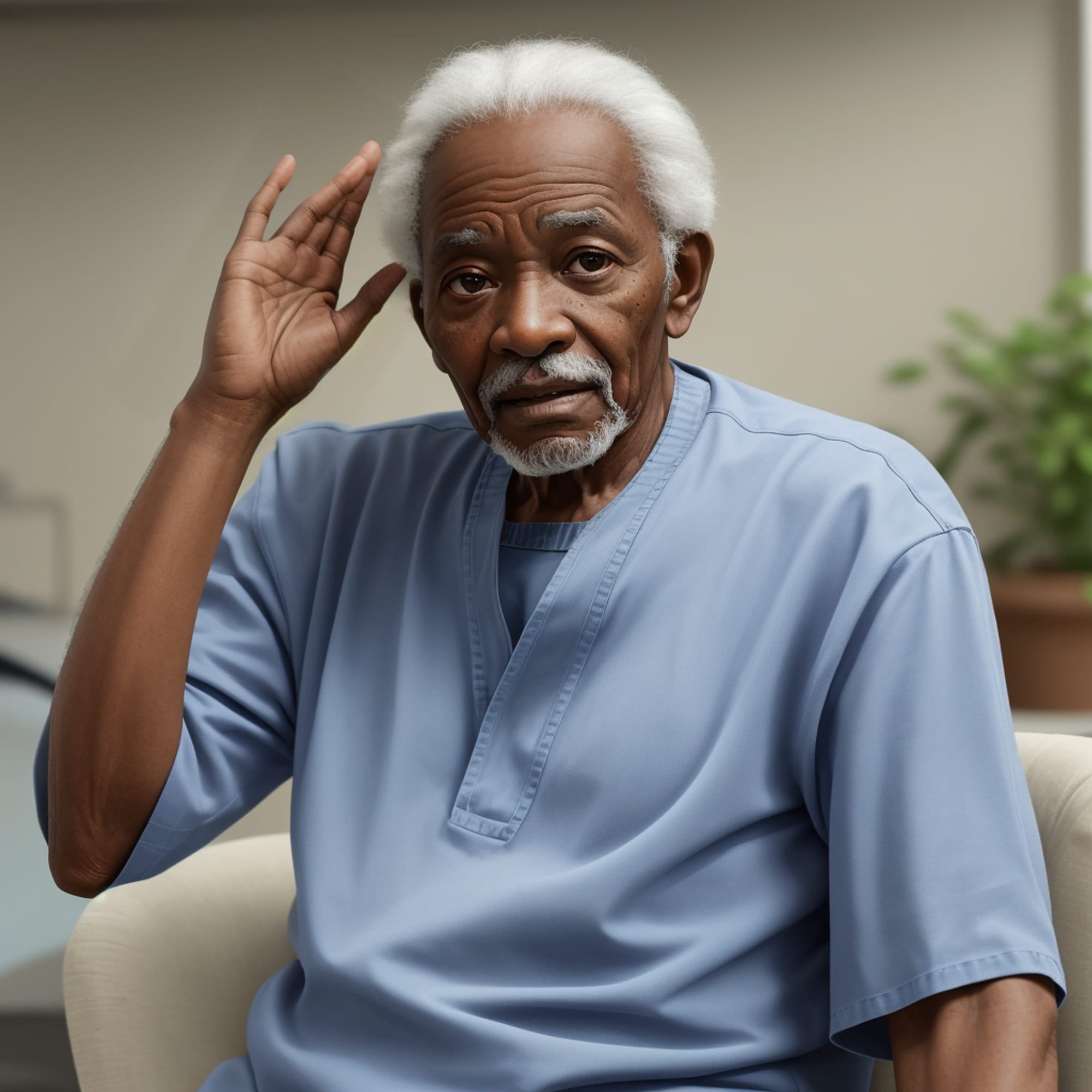}
    \caption*{Restored}
\end{subfigure}%
\vspace{-6pt}
\caption{Silhouette-guided consistent hand restoration.}
\label{fig:hand_poses}
\end{figure}

\subsection{Control Image Generation}
% \subsection{\rev{Conditioning Signals Generation}}
The two main challenges when generating the conditioning signals for hand restoration are (1) ensuring the hand template $T$ is anatomically plausible for the body pose; and (2) accurately positioning $T$ in the input image to make a seamless generation, inclusive of its rotation and handedness (left or right hand). Our detailed solutions to address these two challenges are provided below.

\paragraph{Ensuring Anatomical Plausibility.}
To guarantee the anatomical plausibility of $T$, \rev{we utilize a predefined library~\cite{hand_fix_youtube_template_2023} to 
select an appropriate hand template. }
% While relying on predefined templates might constrain diversity, it
This hand template serves as a novel control image, guiding ControlNet to generate anatomically accurate images. 
This significantly enhances the restoration's anatomical accuracy—a critical factor since methods like mesh fitting (e.g., using MeshFormer~\cite{meshformer_2022}) for severely deformed hands in $M_d$ can lead to unnatural hand shapes that deviate from typical human anatomy. Such deviations are evident when observing inputs with malformed hands in which fingers may appear, \rev{with visualization examples provided in the supplement}. In contrast, the template-based approach of aligning $T$ within the region defined by $M_d$ and subsequently adjusting within the union mask $M = M_d \cup M_t$ ensures a more faithful restoration, demonstrating the method's efficacy in maintaining anatomical fidelity.

\rev{To select a hand template, we propose two methods: (1) silhouette-consistent and (2) random selection. 
Silhouette-consistent selection aims to encourage geometric consistency with the original malformed hand, insofar as is possible while still correcting anatomical issues.
To do so, we generate multiple renders with different hand template parameters and automatically select the render that most closely matches the silhouette, as shown in \cref{fig:hand_poses}.
In contrast, random selection from the template library encourages hand gesture diversity.
The intended use-case for this approach is for a user to select their preferred gesture from a set of renders.
While we do not consider this co-design element in the experiments of this paper, it is part of our creativity-enabling software package.
Note that the primary goal of HandCraft is to ensure anatomical correctness of hand renders and seamless integration with the original image, rather than consistency with the original (potentially corrupt) hand render.
}

\rev{
Finally, we note that artifacts may emerge at the border of the regenerated area. To mitigate these, we generate multiple results where the repainted region has slightly different scales and select the result with the highest realism according to our metrics. This approach ensures that the final output maintains a high level of realism and seamlessly blends with the original image.
}

\begin{figure}[!t]
\centering
% Row 1 images
\begin{subfigure}{.15\textwidth}
  \centering
  \includegraphics[width=\linewidth]{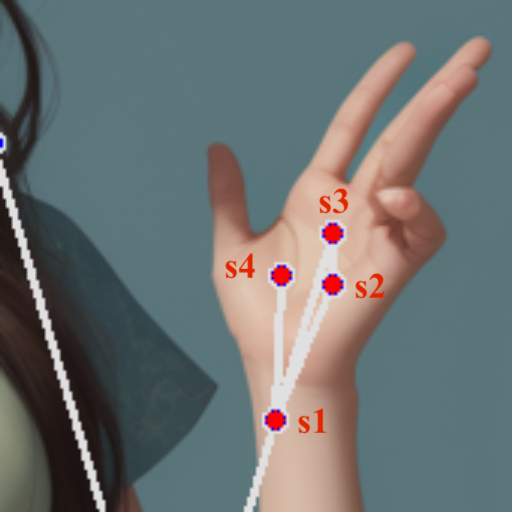}
  \caption*{Detected Keypoints}
\end{subfigure}%
\hfill
\begin{subfigure}{.15\textwidth}
  \centering
  \includegraphics[width=\linewidth]{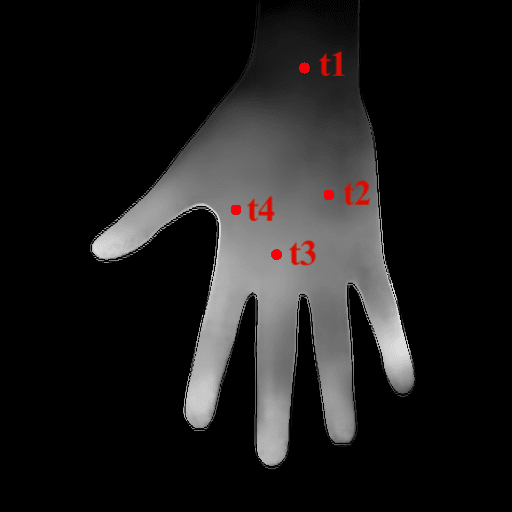}
  \caption*{Alignment Keypoints}
\end{subfigure}%
\hfill
\begin{subfigure}{.15\textwidth}
  \centering
  \includegraphics[width=\linewidth]{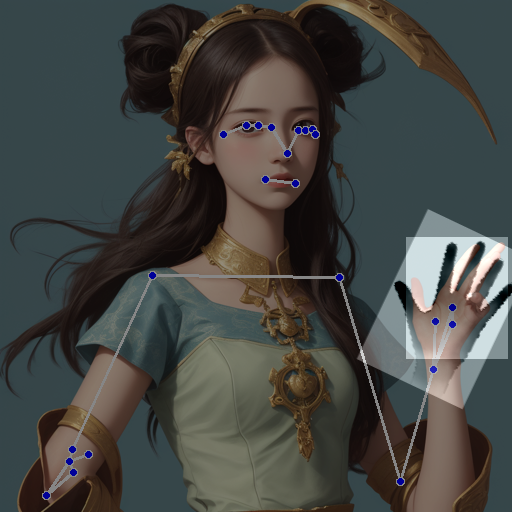}
  \caption*{Overlaid Template}
\end{subfigure}%
\vspace{-4pt}
\caption{\textbf{Alignment of Hand Keypoints.} The left image illustrates a real hand with keypoints $s_1$, $s_2$, $s_3$ and $s_4$ detected by a pose estimation algorithm. These points correspond to critical anatomical landmarks necessary for accurate hand posing. The right image displays a hand template with annotated keypoints $t_1$, $t_2$, $t_3$, and $t_4$, which are intended to align with the keypoints of the real hand after correcting for scale, position, and rotation. The process involves scaling the template based on vector lengths, moving it to match the keypoint positions, and rotating it accordingly.
}
\label{fig:hand_keypoints}
\end{figure}

\begin{figure*}[!t]
\centering
% Row 1 images
\begin{subfigure}{.24\textwidth}
  \centering
  \includegraphics[trim=0 300pt 0 0, clip,width=\linewidth]{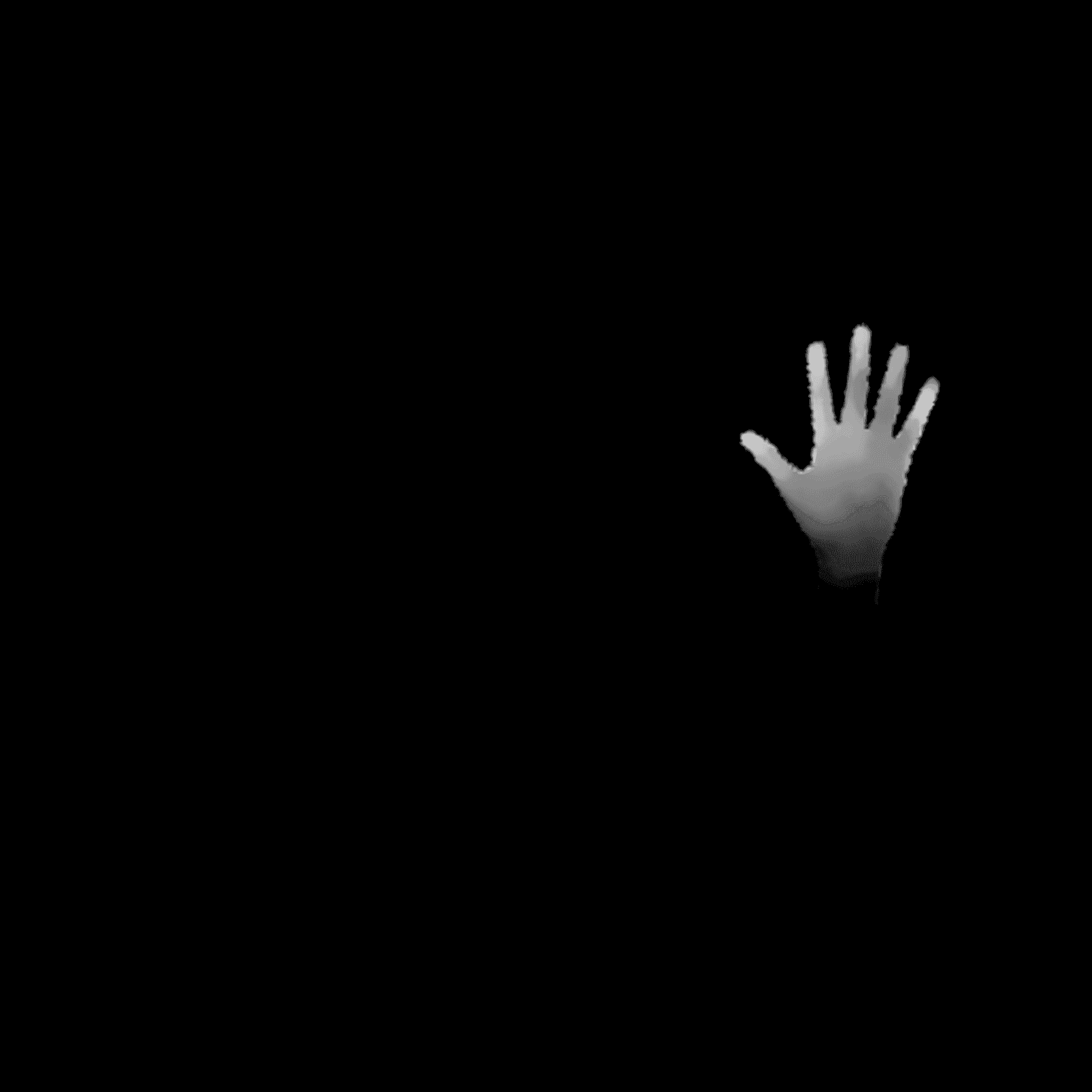}
\end{subfigure}%
\hfill
\begin{subfigure}{.24\textwidth}
  \centering
  \includegraphics[trim=0 300pt 0 0, clip,width=\linewidth]{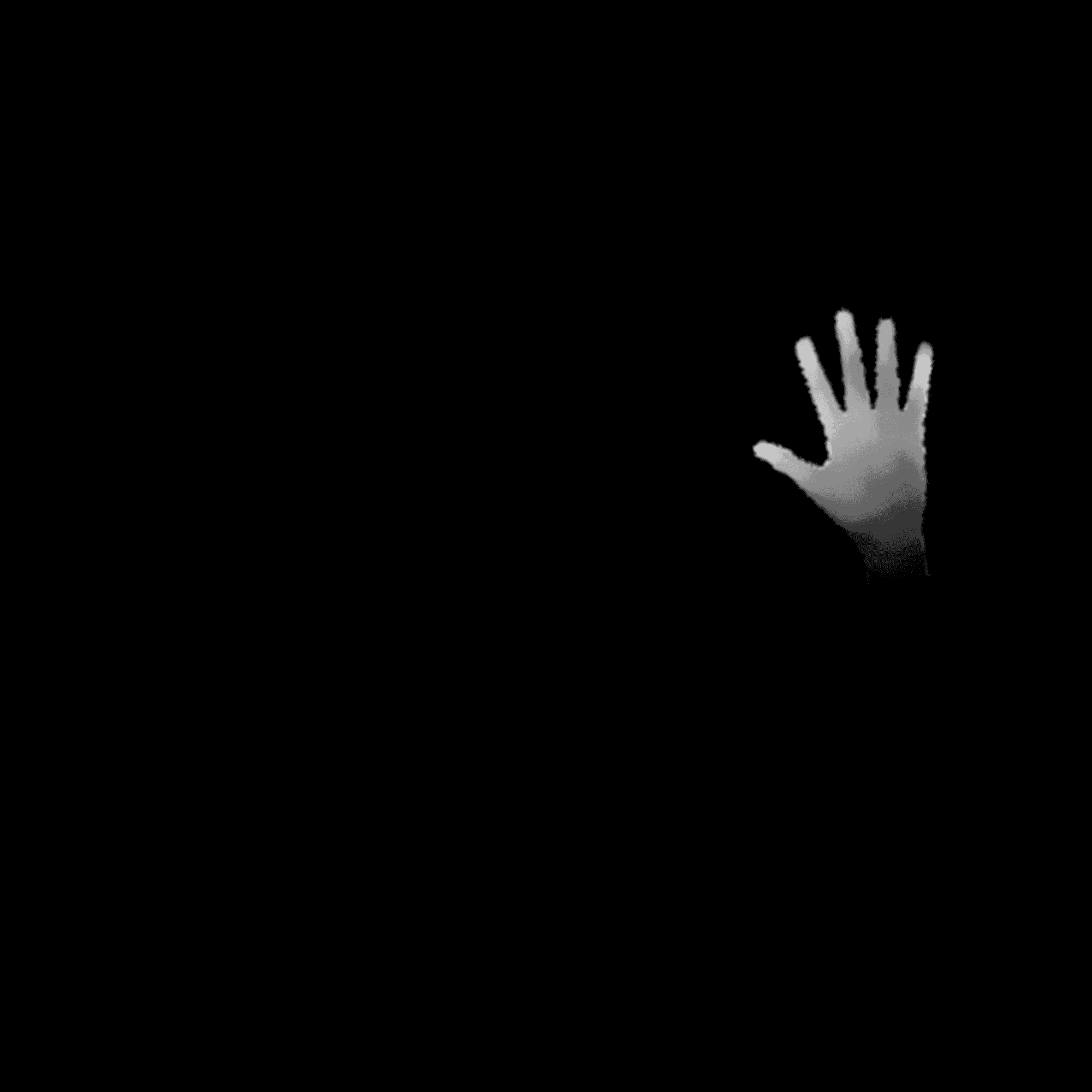}
\end{subfigure}%
\hfill
\begin{subfigure}{.24\textwidth}
  \centering
  \includegraphics[trim=0 300pt 0 0, clip,width=\linewidth]{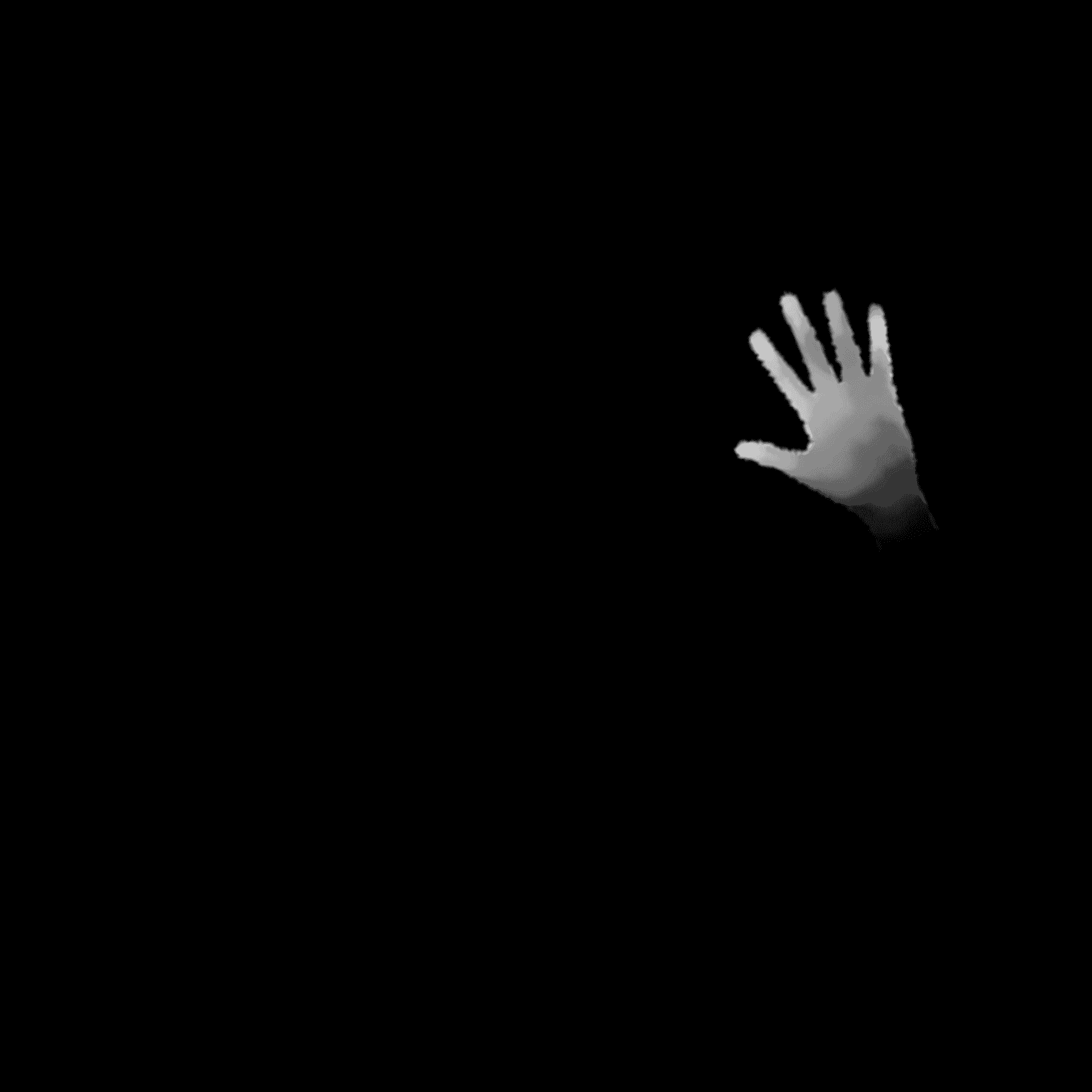}
\end{subfigure}%
\hfill
\begin{subfigure}{.24\textwidth}
  \centering
  \includegraphics[trim=0 300pt 0 0, clip,width=\linewidth]{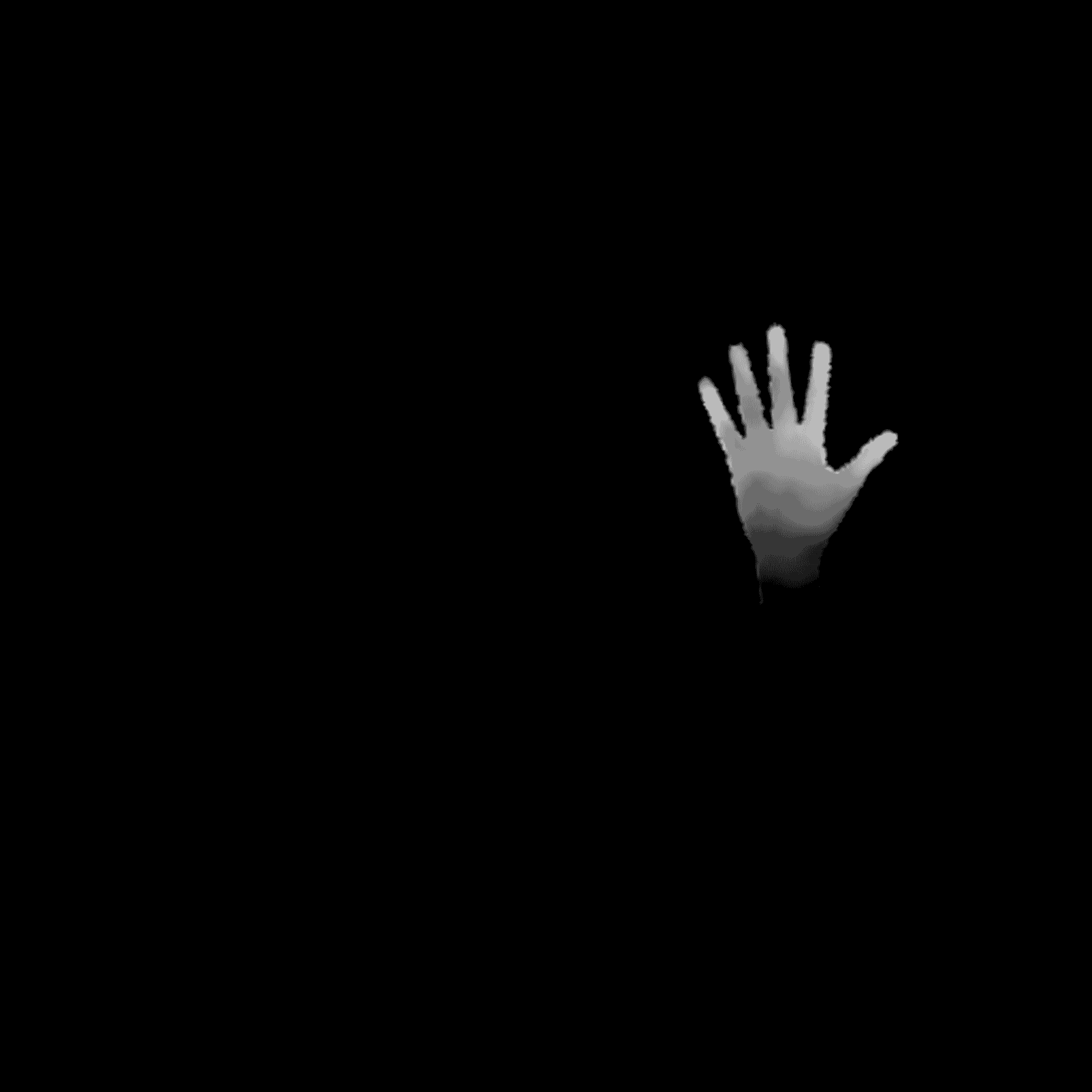}
\end{subfigure}\vfill%
% Row 2 images
\begin{subfigure}{.24\textwidth}
  \centering
  \includegraphics[trim=0 300pt 0 0, clip, width=\linewidth]{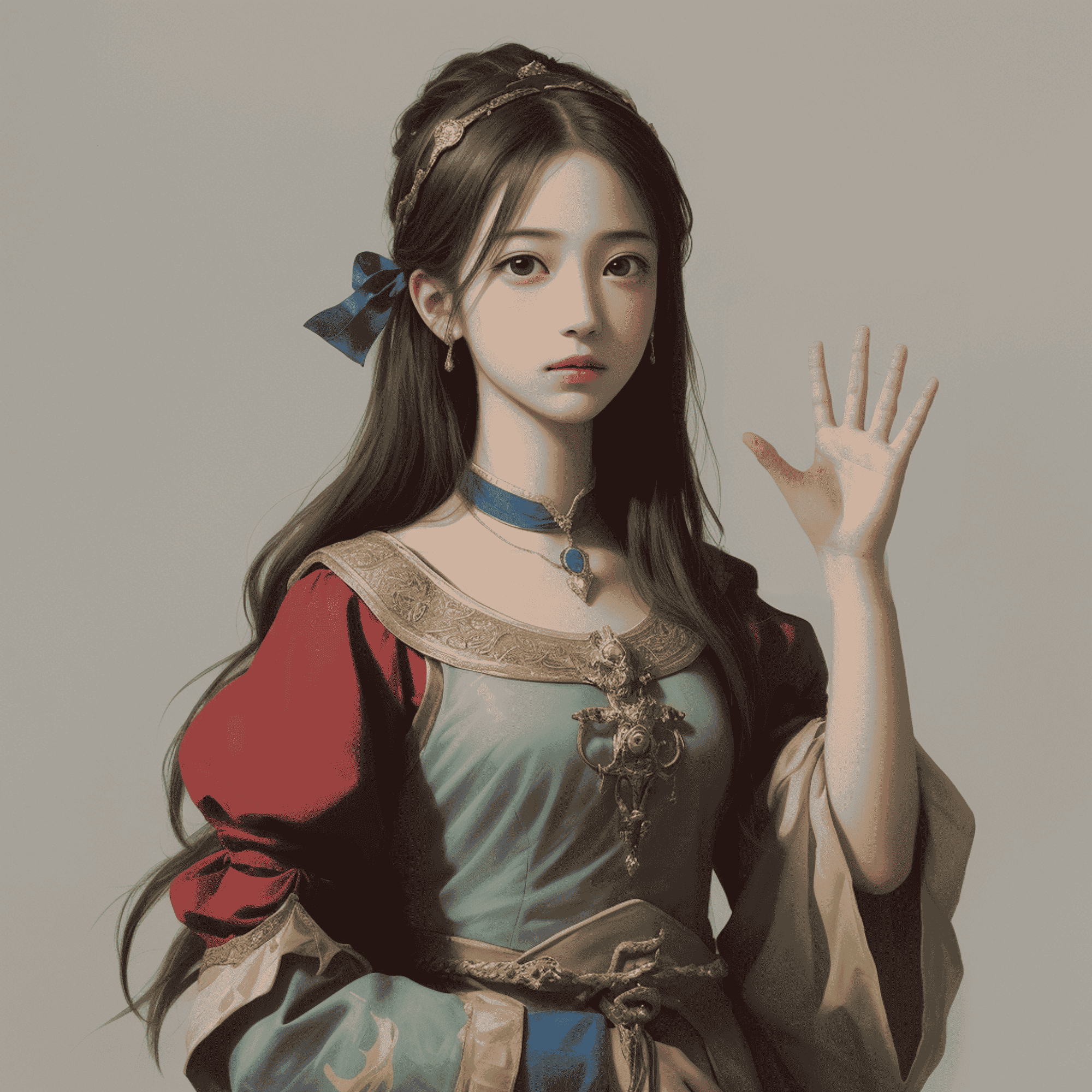}
  \caption{No Rotation}
\end{subfigure}%
\hfill
\begin{subfigure}{.24\textwidth}
  \centering
  \includegraphics[trim=0 300pt 0 0, clip,width=\linewidth]{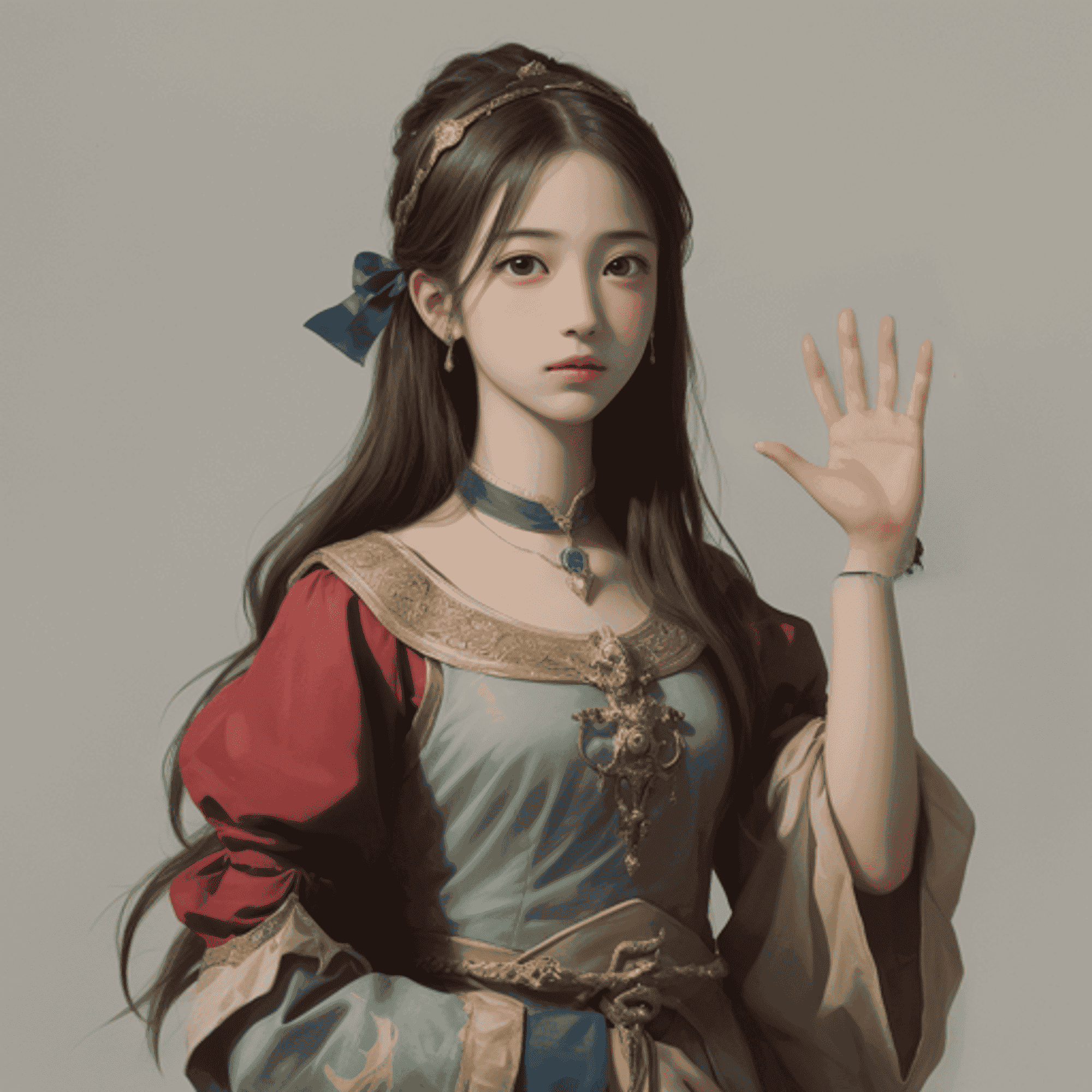}
  \caption{15° Rotation}
\end{subfigure}%
\hfill
\begin{subfigure}{.24\textwidth}
  \centering
  \includegraphics[trim=0 300pt 0 0, clip,width=\linewidth]{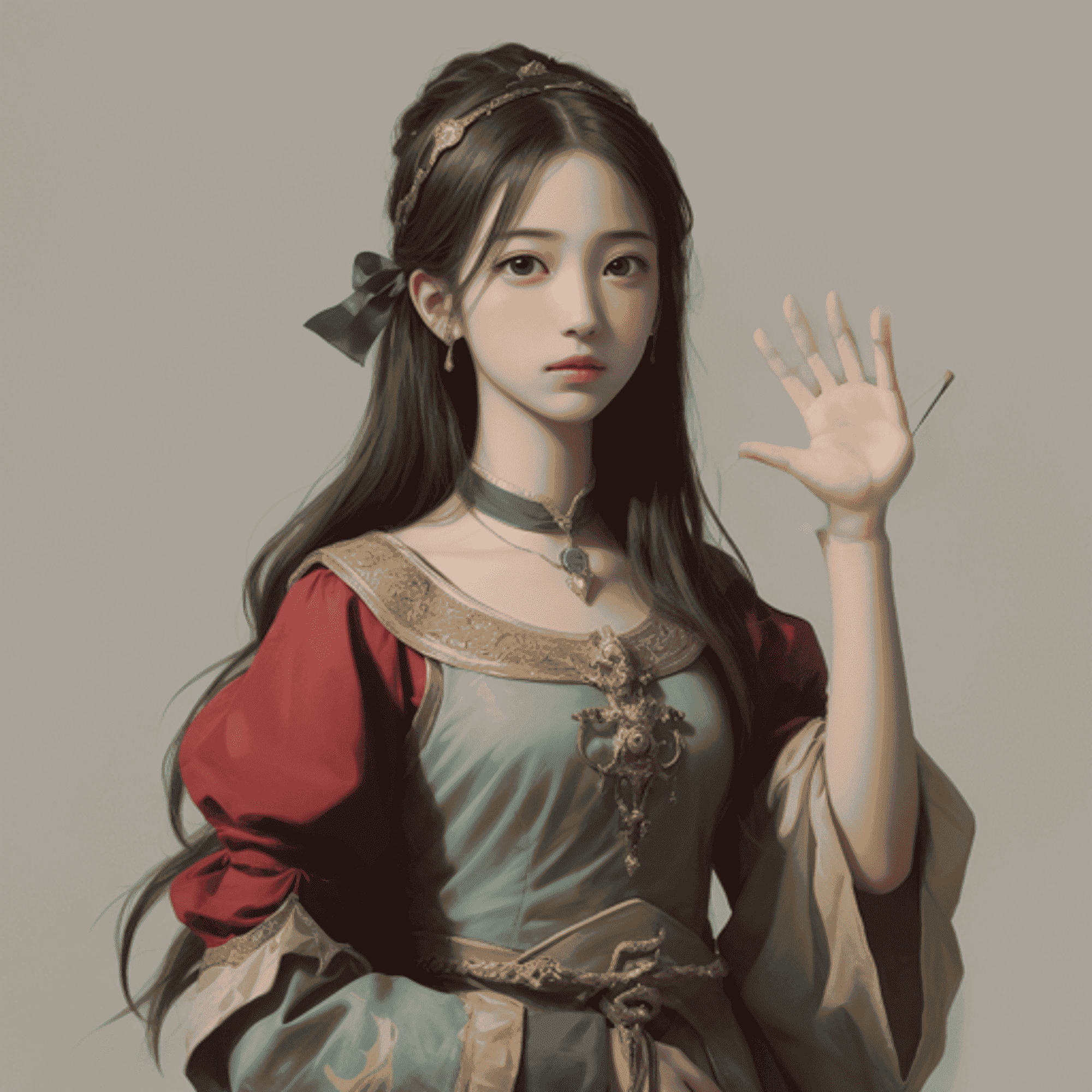}
  \caption{30° Rotation}
\end{subfigure}%
\hfill
\begin{subfigure}{.24\textwidth}
  \centering
  \includegraphics[trim=0 300pt 0 0, clip,width=\linewidth]{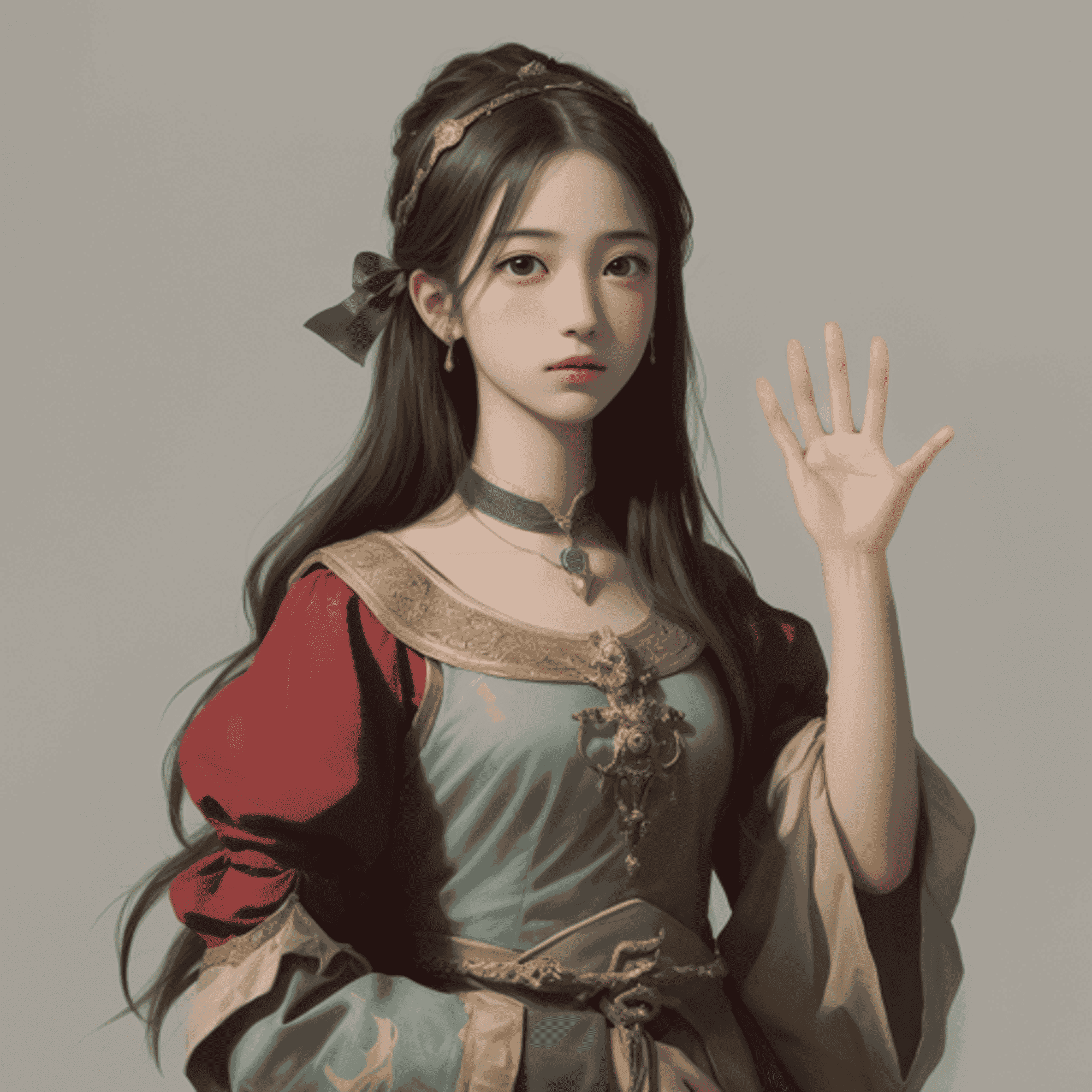}
  \caption{Flipping}
\end{subfigure}%
\vspace{-4pt}
\caption{\textbf{Impact of Template Rotation on Hand Restoration.}
(a) The original hand without any rotation.
(b) The effect of a $15^\circ$ rotation, leading to a misalignment at the wrist and disrupting the natural flow from the forearm to the hand.
(c) The effect of a $30^\circ$ rotation, which further exaggerates the misalignment and creates an unnatural hand shape.
(d) The effect of flipping the hand template, resulting in a mirrored appearance that is anatomically incorrect for that side of the body.
These examples highlight the importance of accurate rotational alignment and proper handedness, where even minor inaccuracies can significantly compromise the restoration's anatomical precision.}
\label{fig:rotation}
\end{figure*}

\paragraph{Accurate Positioning and Orientation.}
The restoration of deformed hands necessitates that the hand is not only anatomically accurate but also precisely positioned and oriented. This means that the hand template $T$ should align correctly in terms of location, rotation, and handedness (left or right hand), corresponding to the detected deformation. \cref{fig:rotation} illustrates that an inaccurate rotation will result in misalignment between the hand and the wrist, and that the wrong handedness (chirality) leads to generation errors.

As shown in \cref{fig:hand_keypoints}, the procedure is as follows.
\begin{enumerate}[nosep]
    \item \textbf{Identification of Keypoints:} Utilizing pose estimation (\eg, MediaPipe~\cite{mediapipe_2019}), we identify keypoints $S_h = \{s_1, s_2, ..., s_n\} \subset S$ on the deformed hand within $M_d$. Corresponding keypoints are also defined on the hand template $T$, denoted as $T_h = \{t_1, t_2, ..., t_n\}$.
    \item \textbf{Scaling:} The \rev{hand} template $T$ is scaled to match the size of the detected hand, based on the ratio of distances between key pairs in $S_h$ and $T_h$, ensuring $T$ fits the size of the deformation in $M_d$.
    \item \textbf{Translation:} $T$ is then translated such that its reference point (e.g., the base of the palm) aligns with the equivalent point in $S$.
    \item \textbf{Rotation:} To correct for orientation, a rotation matrix $R(\theta)$ is applied to $T$, where $\theta$ is the angle calculated from the orientation discrepancy between $S_h$ and $T_h$. For handedness correction, a conditional mirroring transformation may also be applied if necessary.
\end{enumerate}
This ensures that the transformed \rev{hand} template $T$ aligns accurately with the orientation and position of the detected hand within the image $I$.
This procedure highlights the significance of precise alignment in hand restoration, where even minor deviations can lead to an unnatural appearance. By applying these steps, we ensure that the restored hand is correctly aligned and integrated within the input image, maintains anatomical accuracy and seamlessly blends into the original scene, thus preserving the overall authenticity of the image.

\subsection{Hand Restoration}
To ensure that the hand restoration process is targeted and does not inadvertently modify other parts of the image, we introduce the concept of a restoration region. This guides ControlNet~\cite{controlnet_2023} to concentrate exclusively on the identified deformity. Such a strategy reflects our goal of maintaining the original image's integrity, ensuring only the malformed hand is rectified.

The restoration region should:
(1) encompass the entire area of the malformed hand, ensuring comprehensive correction, as represented by the mask $M_d$; and
(2) be sufficiently large to accommodate the corrected hand shape, \ie, the hand template $T$, as represented by the template bounding box mask $M_t$.
This ensures that the restoration does not introduce any spatial constraints that could compromise the correction's effectiveness.
The final restoration region, denoted as $M$, is then given by the union of $M_d$ and $M_t$ ($M = M_d \cup M_t$). This union ensures that the restoration region is optimally sized to cover both the detected deformity and the area required for the corrected hand shape.

This methodical definition of the restoration region is pivotal, as it directly influences the restoration's quality and the preservation of the image's overall composition. Experimental analyses affirm the efficacy of this dual-region approach, highlighting its superiority in achieving precise and aesthetically cohesive hand restorations within the broader context of the original images.

\section{Experiments}

In this section, we present our experiments, where we evaluate the performance of HandCraft qualitatively and quantitatively and compare it to a baseline method.

\subsection{Dataset}
We propose MalHand datasets, including training and evaluation datasets. The evaluation dataset is divided into photorealistic images (MalHand-realistic) and stylized artistic images (MalHand-artistic).
Here, we outline the process of generating these datasets.

\paragraph{Training data.}

While using a pretrained hand detection model, such as YOLOv8 \cite{yolov8}, provides satisfactory performance to detect the malformed hands, greater accuracy can be achieved by finetuning the model.
To do so, it is necessary to compile a training dataset with malformed hands and their locations. For this purpose, we utilize the HaGRID dataset \cite{hagrid_2024}, which contains portrait photos featuring hands, along with bounding boxes that mark the hand positions. 

To create instances of malformed hands for our training data, we leveraged Stable Diffusion \cite{ldm_2022} to modify the hands within the provided bounding boxes, using ``hands'' as the guiding text prompt. This process aimed to generate a variety of hand abnormalities similar to those encountered in images produced by Stable Diffusion models. After generating these modified hands, we manually selected images that clearly displayed malformed hands. This selection process resulted in a dataset comprising 60,000 images, each featuring at least one malformed hand. 

Additionally, to ensure the model can distinguish between malformed and normal hands, we also evaluate our metrics on the unaltered images from HaGRID \cite{hagrid_2024}. The bounding boxes from the original dataset were preserved to provide the locations of both normal and malformed hands.

\paragraph{Evaluation data.}

To assess the effectiveness of our algorithm in restoring malformed hands across various styles, we compiled a dataset comprising 1,500 portrait images featuring malformed hands. This dataset is divided into two categories: 1,000 images in a realistic style (MalHand-realistic) and 500 images in artistic styles (MalHand-artistic). By incorporating a mix of styles, we aim to evaluate the algorithm's robustness and adaptability to different visual representations. The realistic images consist of those sourced from the HaGRID dataset \cite{hagrid_2024}. The generating process is similar to the training data, using a different set of images. These images are used for quantitative evaluation. 

We also generated artistic-style portrait images using Stable Diffusion for qualitative evaluation, including Japanese anime and Disney cartoon styles, among others.
Complete prompts and instructions for generation are provided in the supplement.
We then manually identified portraits with malformed hands. Bounding boxes for these malformed hands were obtained via crowdsourcing, with detailed instructions provided to ensure consistency and accuracy. These instructions are included in the supplement.

\subsection{Evaluation Metrics}

\paragraph{Mean hand pose confidence.}
This metric assesses the naturalness of the hand's anatomy by averaging the confidence scores predicted by Mediapipe~\cite{mediapipe_2019} across all detected hand \textit{keypoints}.
Let $c_{ij}$ denote the confidence of the $j$\textsuperscript{th} hand keypoint out of the set of detected keypoint indices $\mathcal{J}_i$ for hand $i$ in a dataset containing $N$ hands.
Then the mean hand pose confidence is given by
\begin{equation}
\label{eq:cpose}
\bar{c}_{\text{pose}} = \frac{1}{N} \sum_{i=1}^{N} \frac{1}{|\mathcal{J}_i|} \sum_{j \in \mathcal{J}_i} c_{ij}.
\end{equation}

\paragraph{Mean hand classifier confidence.}
This metric assesses the model's ability to generate hands that can be confidently classified as non-malformed by our YOLOv8-based hand detector~\cite{yolov8}.
Let $c_{i}'$ denote the confidence of the hand classifier for hand $i$ in a dataset containing $N$ hands.
Then the mean hand classifier confidence is given by 
\begin{equation}
\label{eq:cclassifier}
\bar{c}_{\text{classifier}} = \frac{1}{N} \sum_{i=1}^{N} c_{i}'.
\end{equation}

\paragraph{Masked peak signal-to-noise ratio (PSNR) / masked structural similarity index measure (SSIM).}
These metrics assess how well the image outside the restored area is preserved, at a per-pixel and structural level: whether the restoration has altered or corrupted the rest of the image.

\begin{figure}[!t]
\centering
\begin{subfigure}[b]{0.48\linewidth}
    \includegraphics[width=\linewidth]{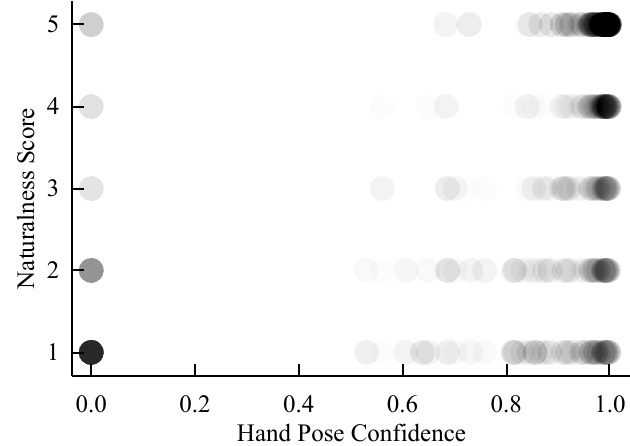}
\end{subfigure}\hfill
\begin{subfigure}[b]{0.48\linewidth}
    \includegraphics[width=\linewidth]{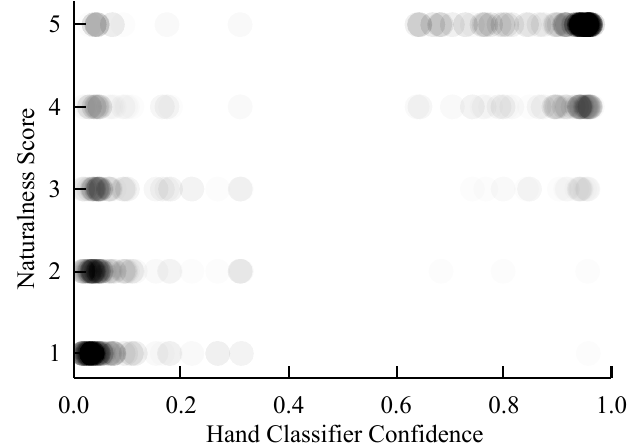}
\end{subfigure}%
\vspace{-6pt}
\caption{Human hand naturalness score has positive correlation with $c_\text{pose}$ and $c_\text{classifier}$}
\label{fig:metrics}
\end{figure}

\paragraph{Validation.}
User studies were conducted to investigate whether the confidence scores (\ie, ${c}_{\text{pose}}$ and ${c}_{\text{classifier}}$) correlated with naturalness.
Seven unaffiliated individuals rated the naturalness of hands in each image, from a stratified random subset of 200 restored images, on a scale of 1 to 5, where 1 is least natural and 5 is most natural.
As shown in \cref{fig:metrics} (opacity 1\%), the hand pose confidence score ($c_\text{pose}$) and the hand classifier confidence score ($c_\text{classifier}$) correlates with perceived naturalness.
The average Pearson's correlation coefficient between $c_\text{pose}$ and human-rated naturalness scores is 0.44 and the average correlation for $c_\text{classifier}$ is 0.82.

\begin{figure}[!t]
\centering
% Row 1 images
\begin{subfigure}{.24\textwidth}
  \centering
  \includegraphics[trim=0 0 0 0, clip, width=\linewidth]{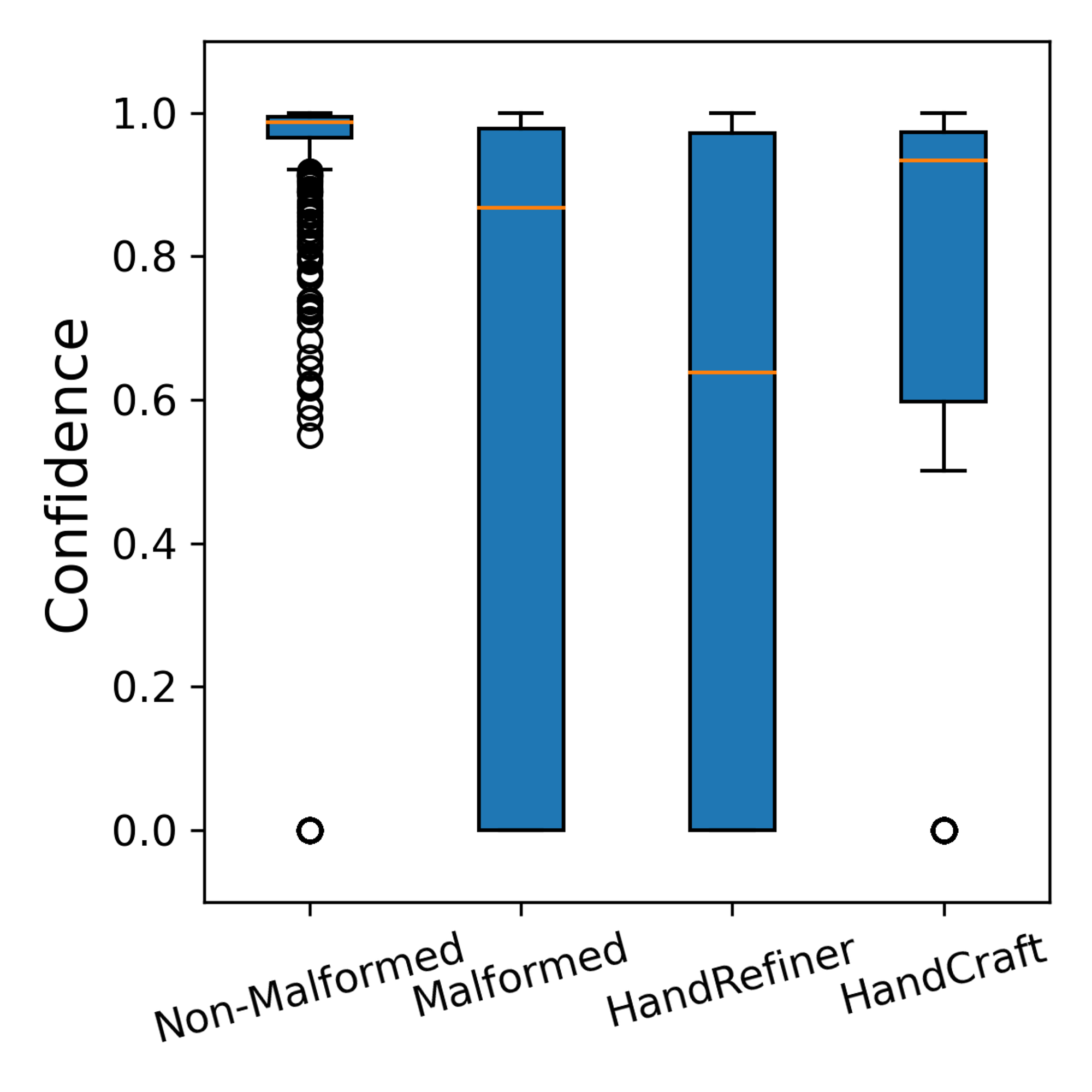}
  \caption{$c_{\text{pose}}$}
\end{subfigure}%
\begin{subfigure}{.24\textwidth}
  \centering
  \includegraphics[trim=0 0 0 0, clip, width=\linewidth]{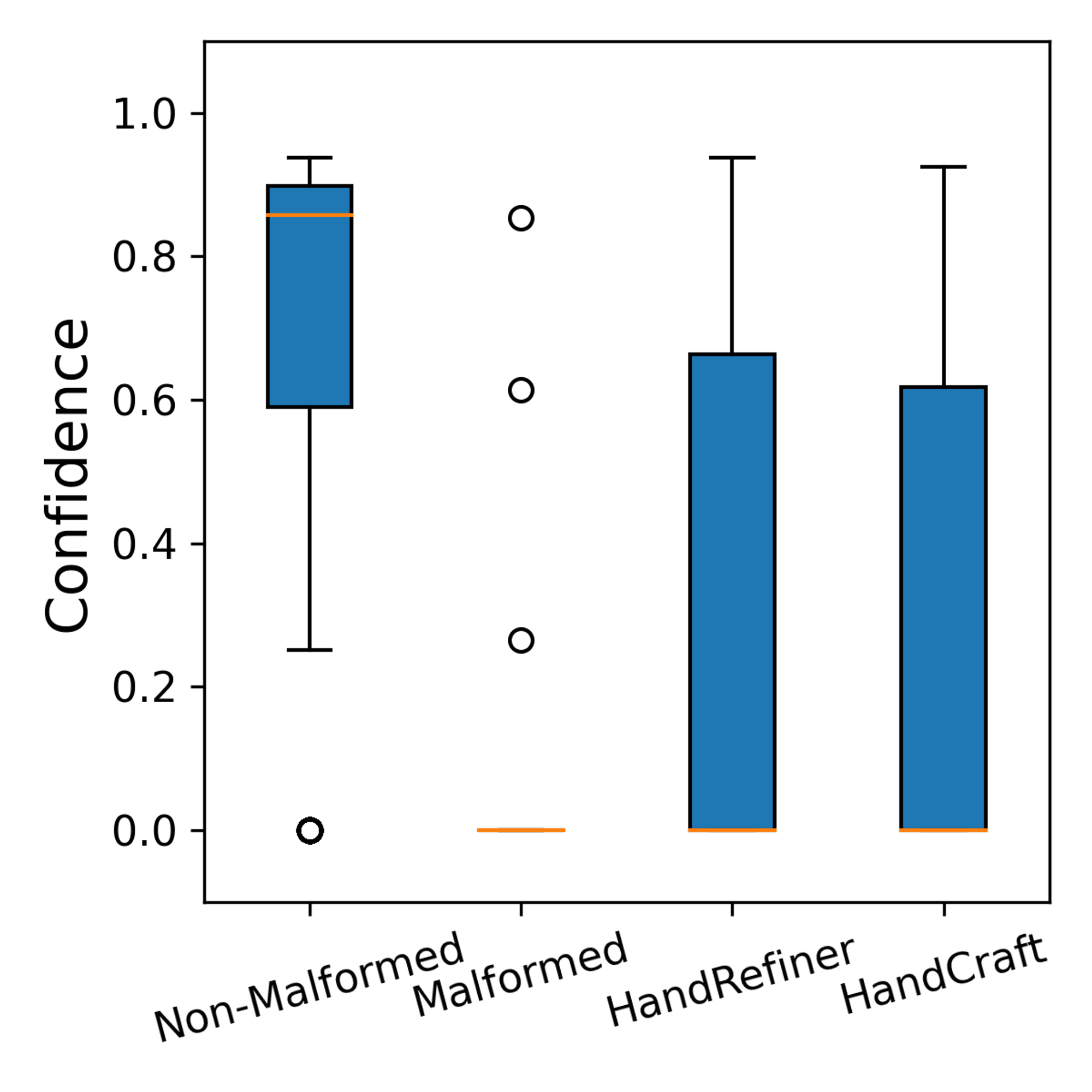}
  \caption{$c_{\text{classifer}}$}
\end{subfigure}%
\vspace{-6pt}
\caption{Box plots demonstrating the performance of different methods in restoring anatomical correctness to images with malformed hands. (a) The hand pose confidence $c_{\text{pose}}$, where HandCraft shows a notable improvement over HandRefiner, closely aligning with the non-malformed control group. (b) The hand classifier confidence $c_{\text{classifier}}$, where HandCraft's restorations achieve comparable confidence levels to HandRefiner.
}
\label{fig:box_plot}
\end{figure}

\subsection{Comparison with \rev{Non-Malformed} Images}
We quantitatively compare the anatomical accuracy of HandCraft's restored images to a control group of images without malformed hands. Two sets are assembled for evaluation: a control set $\mathcal{D}_{\text{C}}$ composed of the real images from the HaGRID dataset~\cite{hagrid_2024} and a set $\mathcal{D}_{\text{R}}$ composed of realistic images with malformed hands from the MalHand-realistic dataset, which have undergone hand restoration with HandCraft.
For each image, we calculate the hand pose confidence ${c}_{\text{pose}}$ and hand classifier confidence ${c}_{\text{classifer}}$, reflecting anatomical correctness as perceived by hand detection and pose estimation algorithms, with higher scores indicating closer resemblance to authentic hand anatomy. 

\cref{fig:box_plot} shows overlapping hand pose confidence intervals between the restored and Non-Malformed images. While there is a statistically significant difference between the two groups, this is partly due to inherent differences between real and restored generated images. The overlap in mean hand pose confidence scores between HandCraft restorations and the Non-Malformed group indicates HandCraft's superior performance in restoring anatomical correctness compared to HandRefiner.

\begin{table}[!t]\centering
\setlength{\tabcolsep}{2pt}
\caption{
\rev{
Quantitative comparison of hand restoration methods on the MalHand-realistic dataset.
We report the mean hand pose confidence ($\bar{c}_{\text{pose}}$) to assess the accuracy of the hand restoration, the mean hand classifier confidence ($\bar{c}_{\text{classifier}}$) to assess whether a classifier deems the hand as non-malformed, the masked peak signal-to-noise ratio (PSNR) to assess the visual fidelity outside of the hand region, and the masked structural similarity index measure (SSIM) to assess any change in structural content.
The latter two are calculated between the input image $I$ with malformed hands and restored images. CtrlNet denotes ControlNet, HP denotes hand pose, and Seg denotes segmentation. 
}
}
\label{tab:sota_comparison}
\begin{tabularx}{\linewidth}{l C C C C}
\toprule
Method & $\bar{c}_\text{pose} \uparrow$ &  $\bar{c}_\text{classifier} \uparrow$ & PSNR $\uparrow$ & SSIM $\uparrow$ \\
\midrule
Null Intervention & 0.68 & 0.00 & N/A & N/A \\
CtrlNetHP~\cite{controlnet_2023} & 0.57 & 0.09 & 12.26 & 0.5848 \\
CtrlNetSeg~\cite{controlnet_2023} & 0.61 & 0.17 & 14.62 & 0.6261 \\
HandRefiner~\cite{handrefiner_2023} & 0.54 & {\textbf{0.25}} & 12.93 & 0.3839 \\
HandCraft (Ours) & {\textbf{0.79}} & {\textbf{0.25}} & {\textbf{23.40}} & {\textbf{0.6462}} \\
\bottomrule
\end{tabularx}
\end{table}

\rev{
\subsection{Comparison with Baselines}
}

\begin{figure*}[t]
\centering
% Row 1
\begin{subfigure}{.05\textwidth}
  \centering
  \raisebox{.11\textheight}{%
    \parbox[c][.23\textheight][c]{1em}{%
      \centering
      \rotatebox{90}{Malformed Hand}%
    }%
  }
\end{subfigure}%
\hfill
\begin{subfigure}{.23\textwidth}
  \centering
  \includegraphics[width=1.0\linewidth]{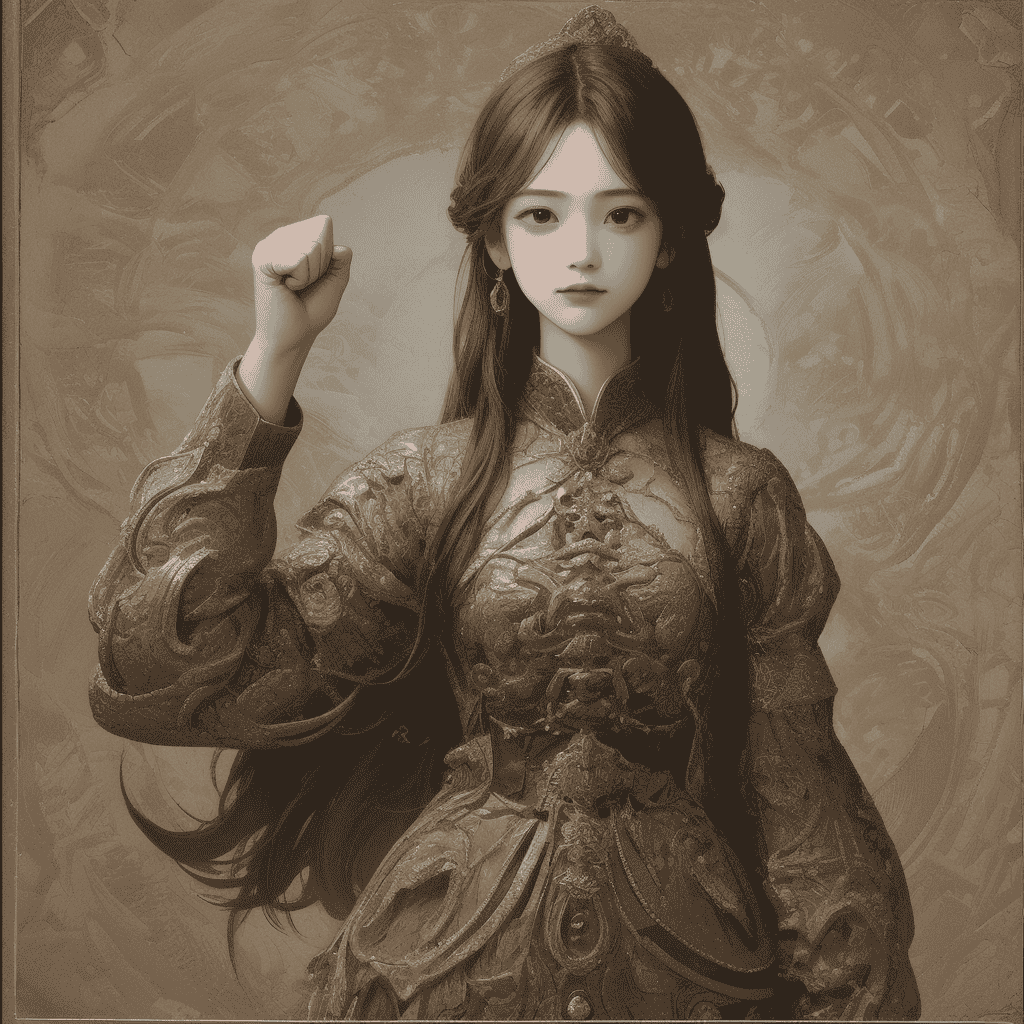}
  \caption*{\rev{Input Image}}
\end{subfigure}%
\hfill
\begin{subfigure}{.23\textwidth}
  \centering
  \includegraphics[width=1.0\linewidth]{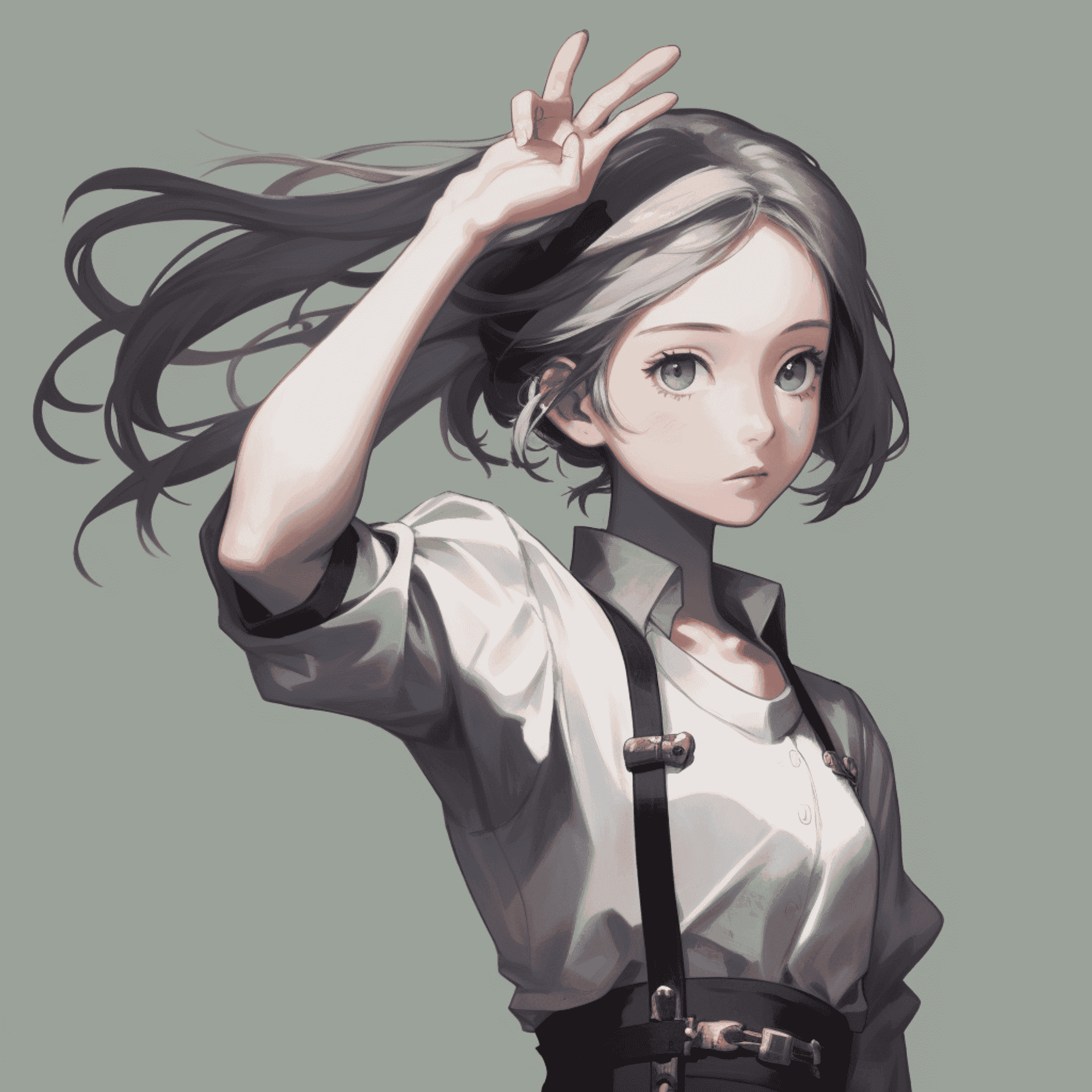}
  \caption*{Input Image}
\end{subfigure}%
\hfill
\begin{subfigure}{.23\textwidth}
  \centering
  \includegraphics[width=1.0\linewidth]{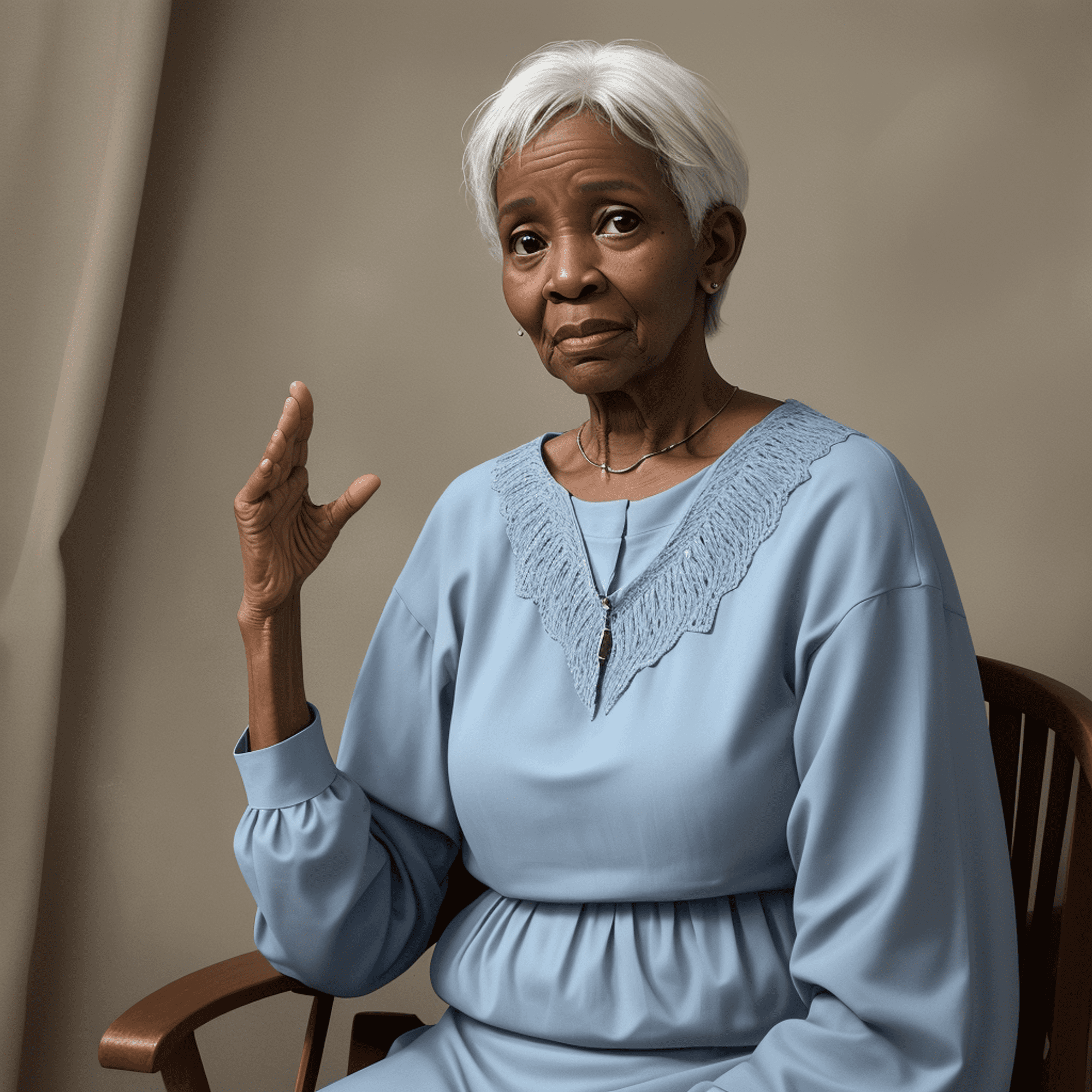}
  \caption*{\rev{Input Image}}
\end{subfigure}
\hfill
\begin{subfigure}{.23\textwidth}
  \centering
  \includegraphics[width=1.0\linewidth]{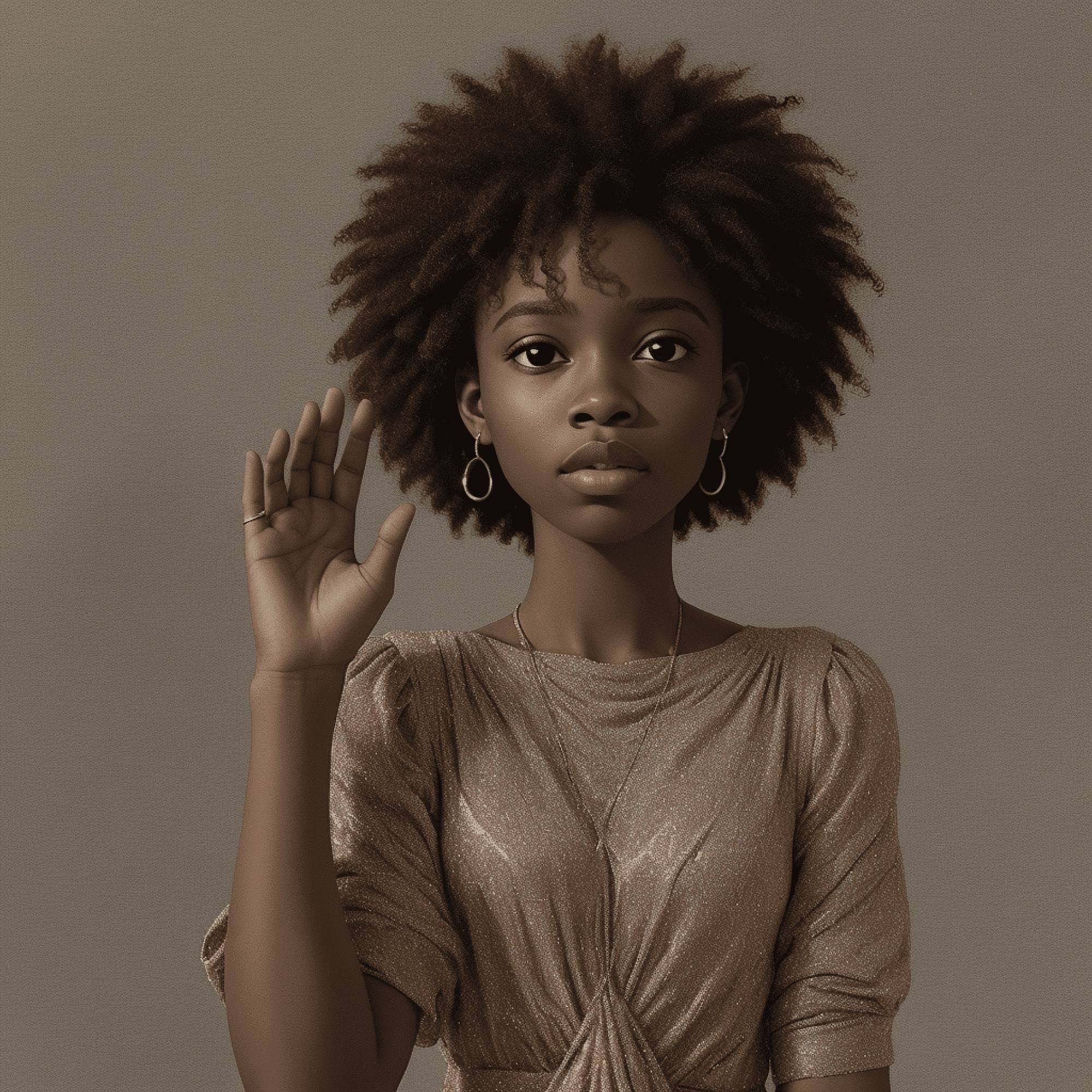}
  \caption*{\rev{Input Image}}
\end{subfigure}
\vspace{-6mm}

% Row 2
\begin{subfigure}{.05\textwidth}
  \centering
  \raisebox{.11\textheight}{%
    \parbox[c][.23\textheight][c]{1em}{%
      \centering
      \rotatebox{90}{HandRefiner~\cite{handrefiner_2023}}%
    }%
  }
\end{subfigure}%
\hfill
\begin{subfigure}{.23\textwidth}
  \centering
  \includegraphics[width=1.0\linewidth]{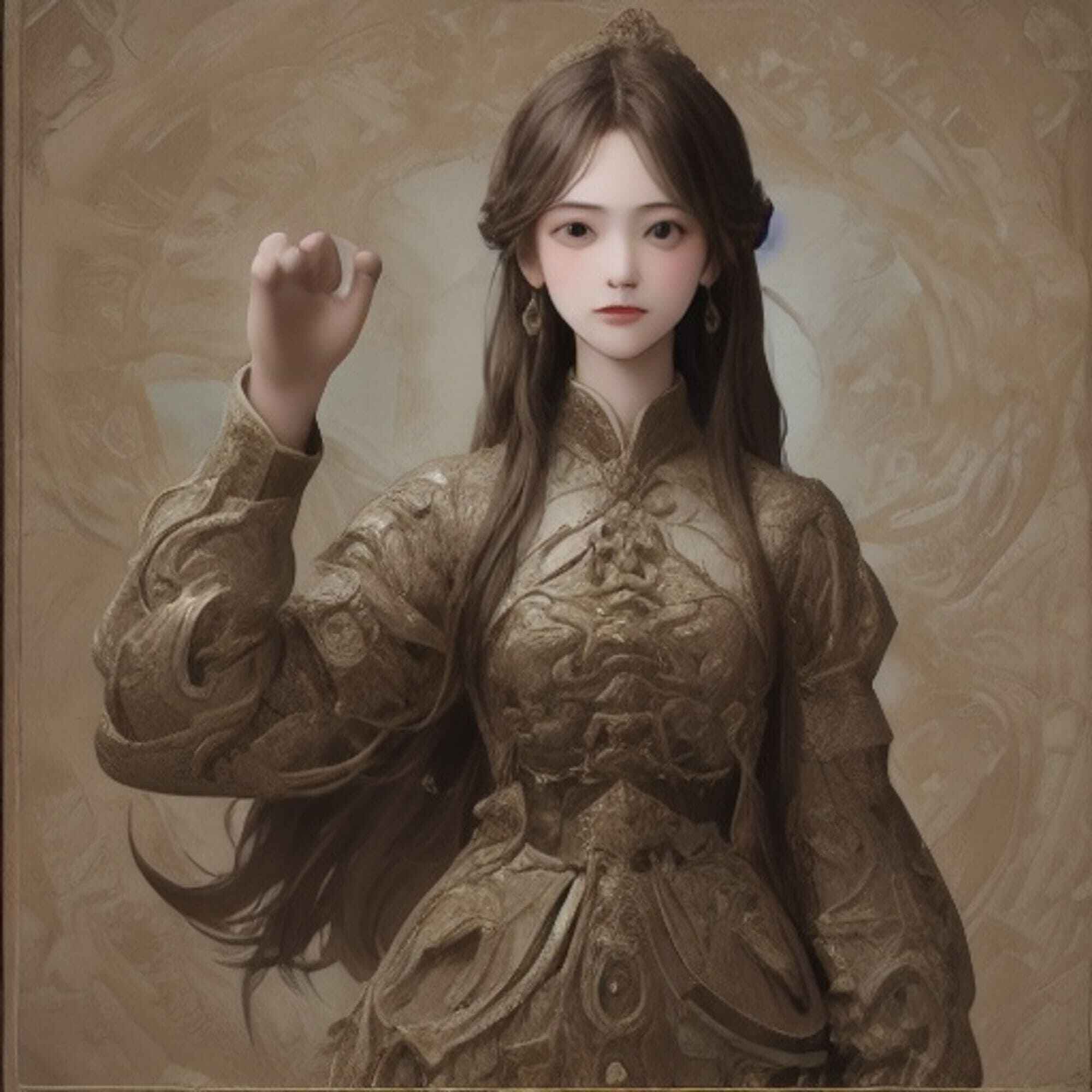}
  \caption*{\rev{PSNR: 28.81 dB / SSIM: 0.7378}}
\end{subfigure}%
\hfill
\begin{subfigure}{.23\textwidth}
  \centering
  \includegraphics[width=1.0\linewidth]{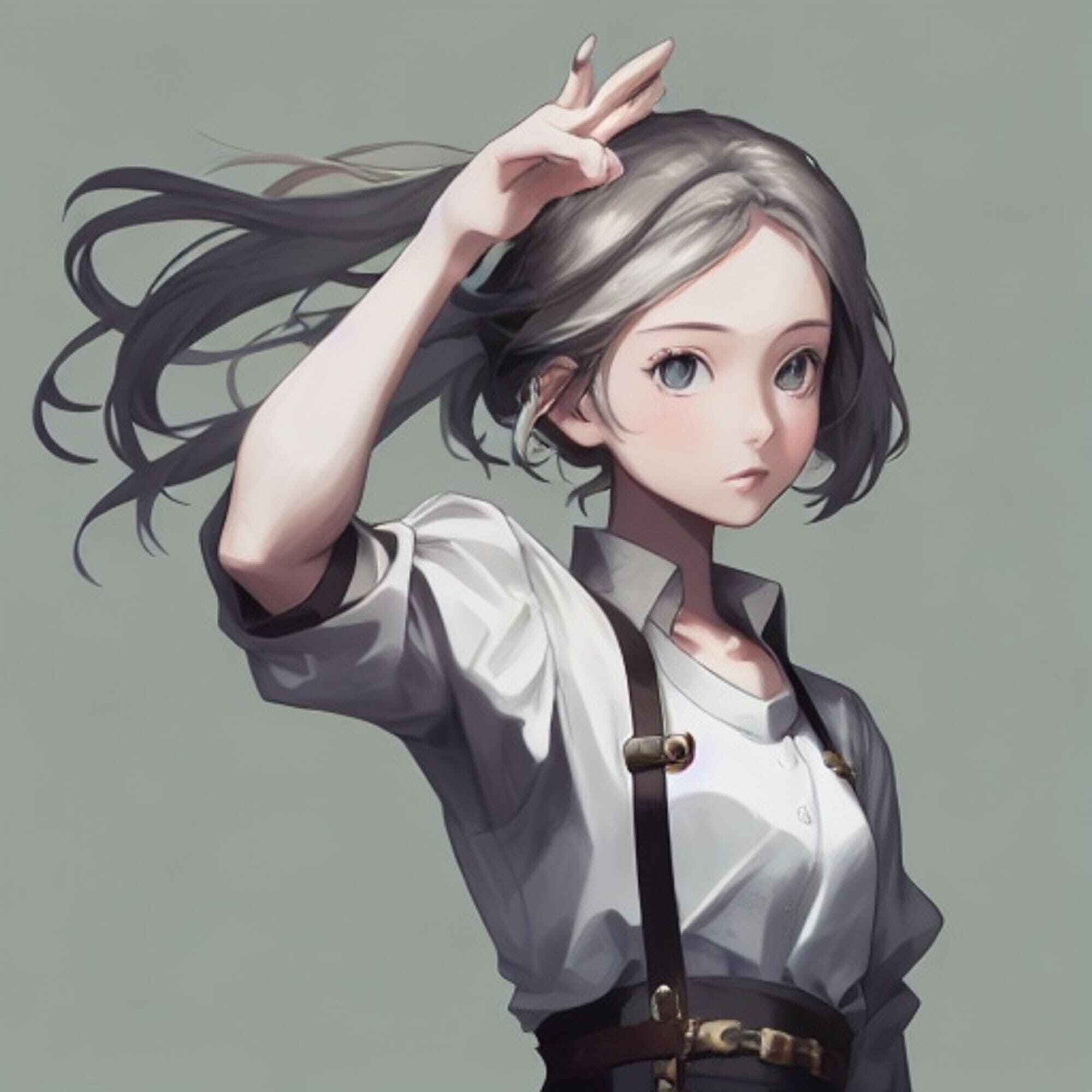}
  \caption*{PSNR: 32.30 dB / SSIM: 0.9510}
\end{subfigure}%
\hfill
\begin{subfigure}{.23\textwidth}
  \centering
  \includegraphics[width=1.0\linewidth]{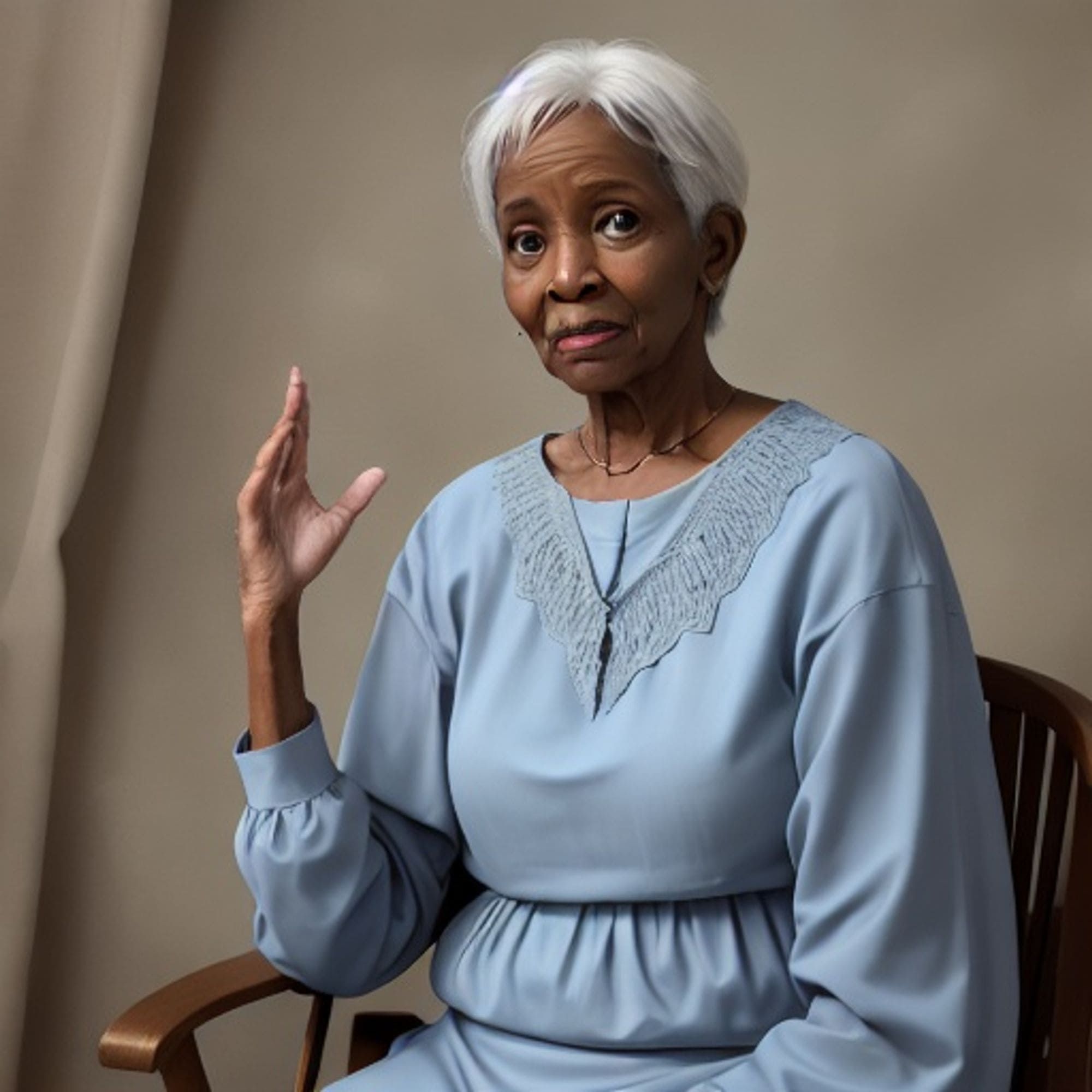}
  \caption*{\rev{PSNR: 27.16 dB / SSIM: 0.7911}}
\end{subfigure}
\hfill
\begin{subfigure}{.23\textwidth}
  \centering
  \includegraphics[width=1.0\linewidth]{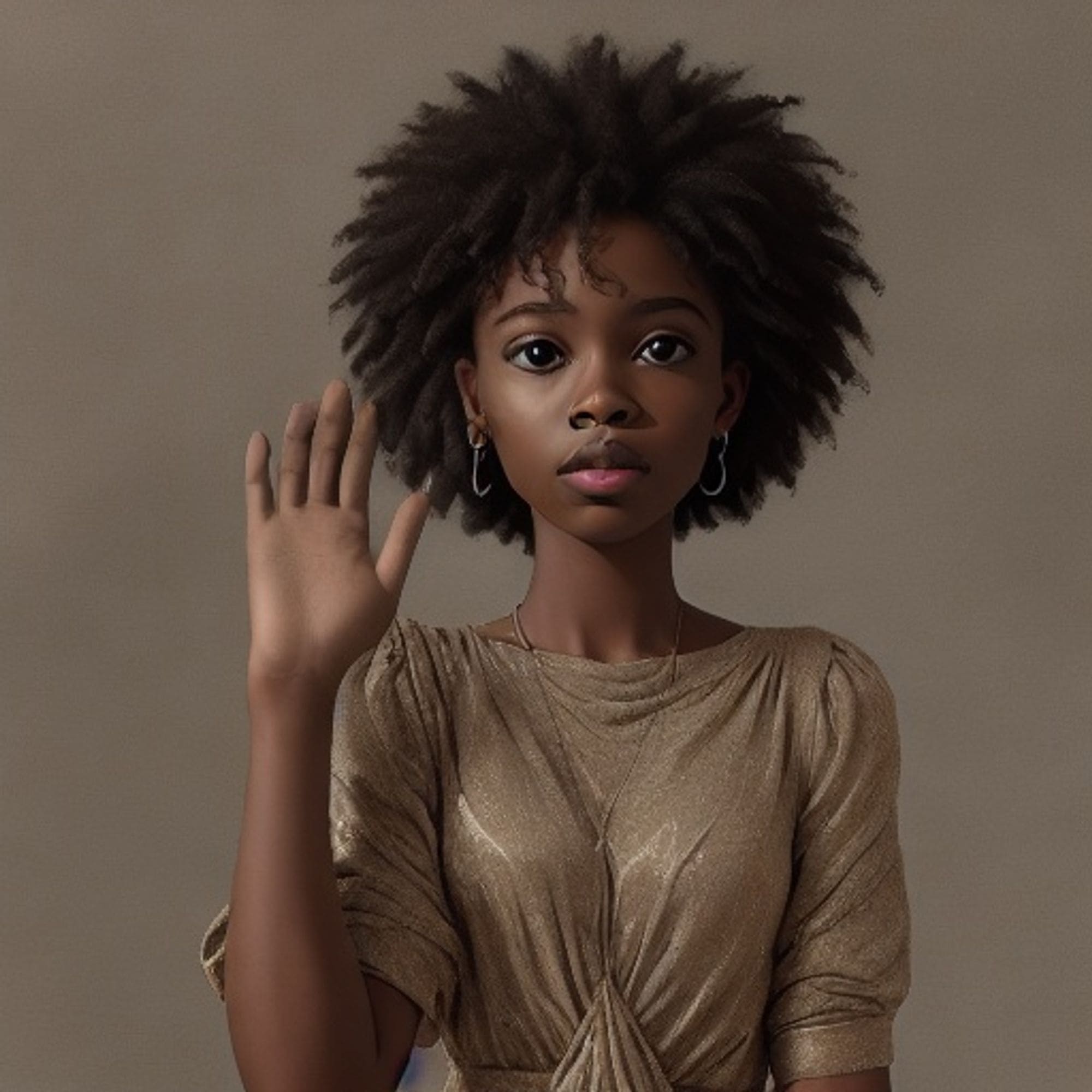}
  \caption*{\rev{PSNR: 27.40 dB / SSIM: 0.6075}}
\end{subfigure}
\vspace{-6mm}

% Row 3
\begin{subfigure}{.05\textwidth}
  \centering
  \raisebox{.11\textheight}{%
    \parbox[c][.23\textheight][c]{1em}{%
      \centering
      \rotatebox{90}{HandCraft}%
    }%
  }
\end{subfigure}%
\hfill
\begin{subfigure}{.23\textwidth}
  \centering
  \includegraphics[width=1.0\linewidth]{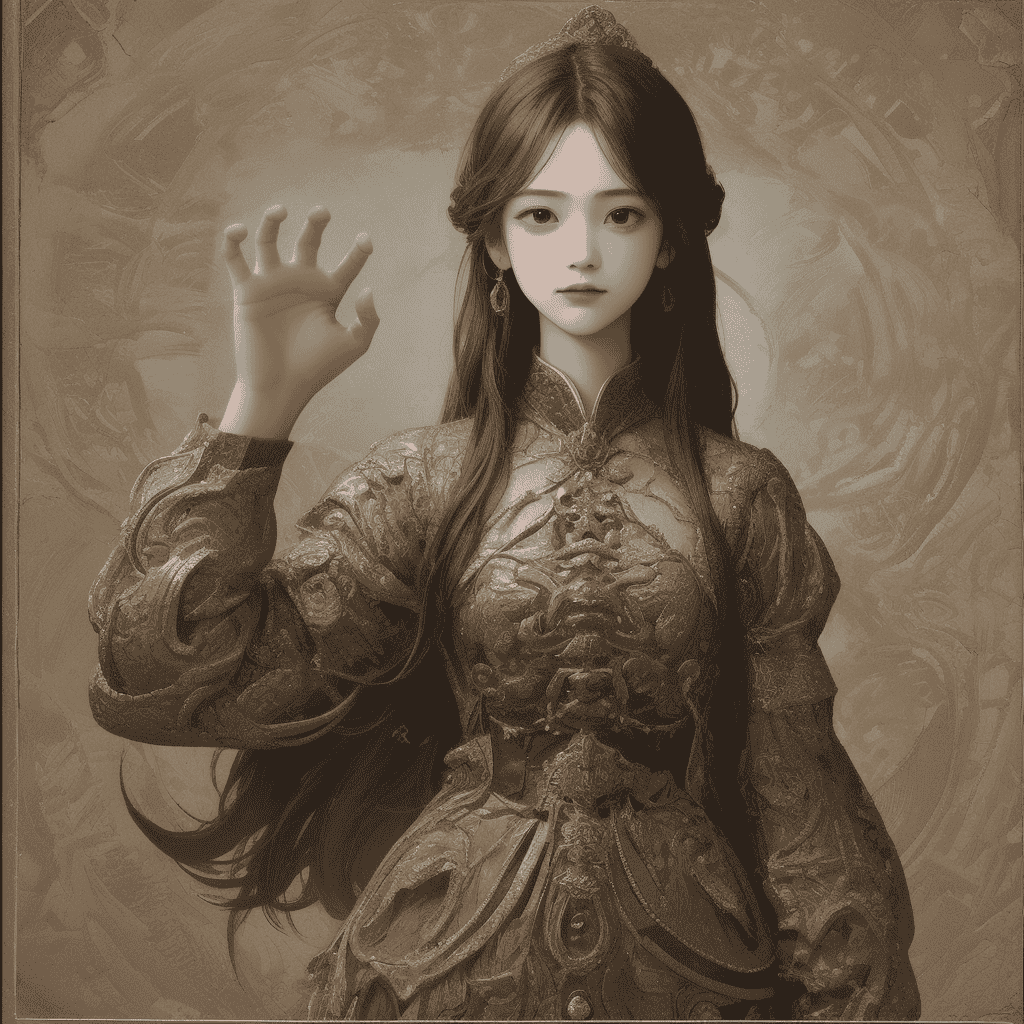}
  \caption*{\rev{PSNR: 42.76 dB / SSIM: 0.9973}}
\end{subfigure}%
\hfill
\begin{subfigure}{.23\textwidth}
  \centering
  \includegraphics[width=1.0\linewidth]{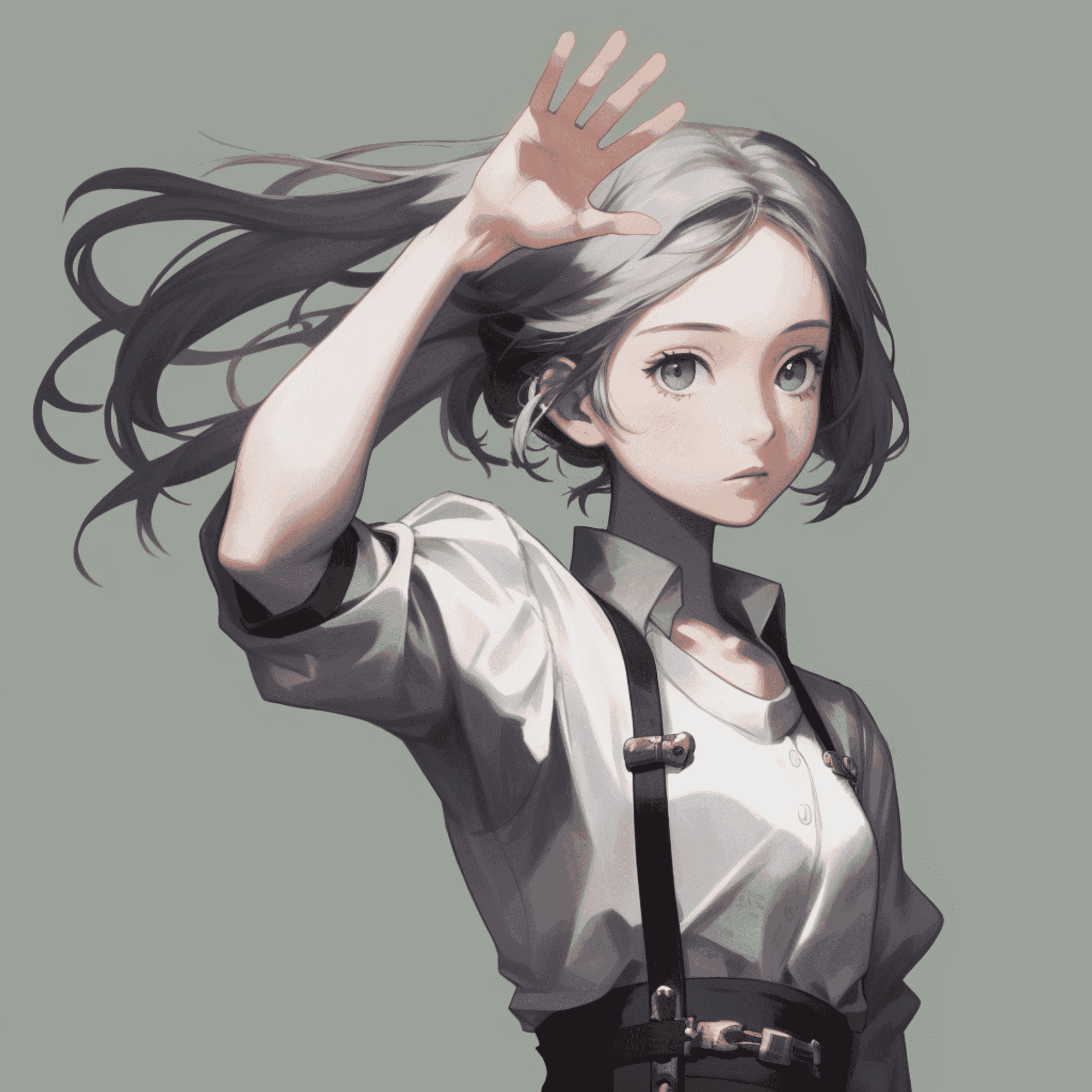}
  \caption*{PSNR:112.9 dB / SSIM: 0.9999}
\end{subfigure}%
\hfill
\begin{subfigure}{.23\textwidth}
  \centering
  \includegraphics[width=1.0\linewidth]{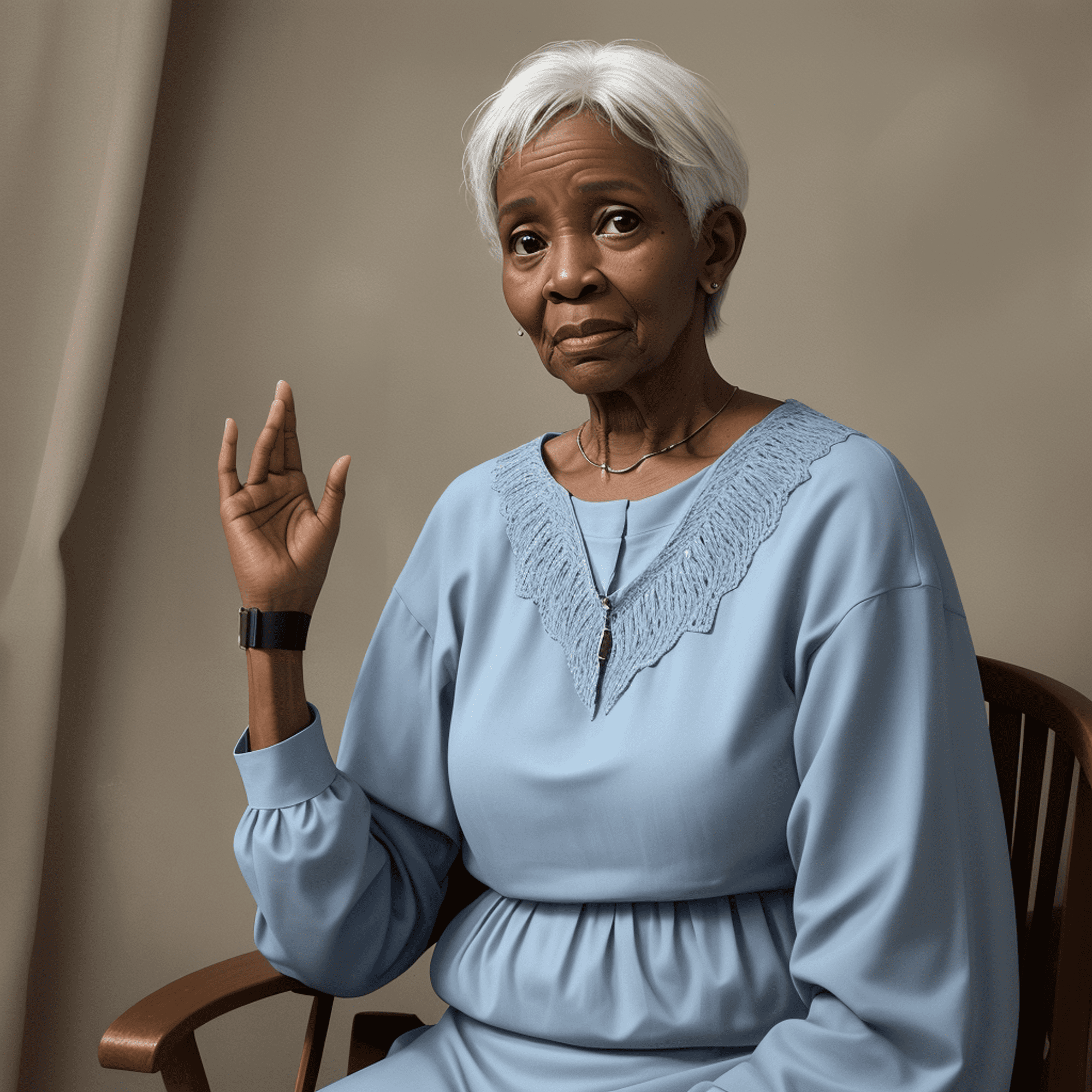}
  \caption*{\rev{PSNR: 29.64 dB / SSIM: 0.8366}}
\end{subfigure}
\hfill
\begin{subfigure}{.23\textwidth}
  \centering
  \includegraphics[width=1.0\linewidth]{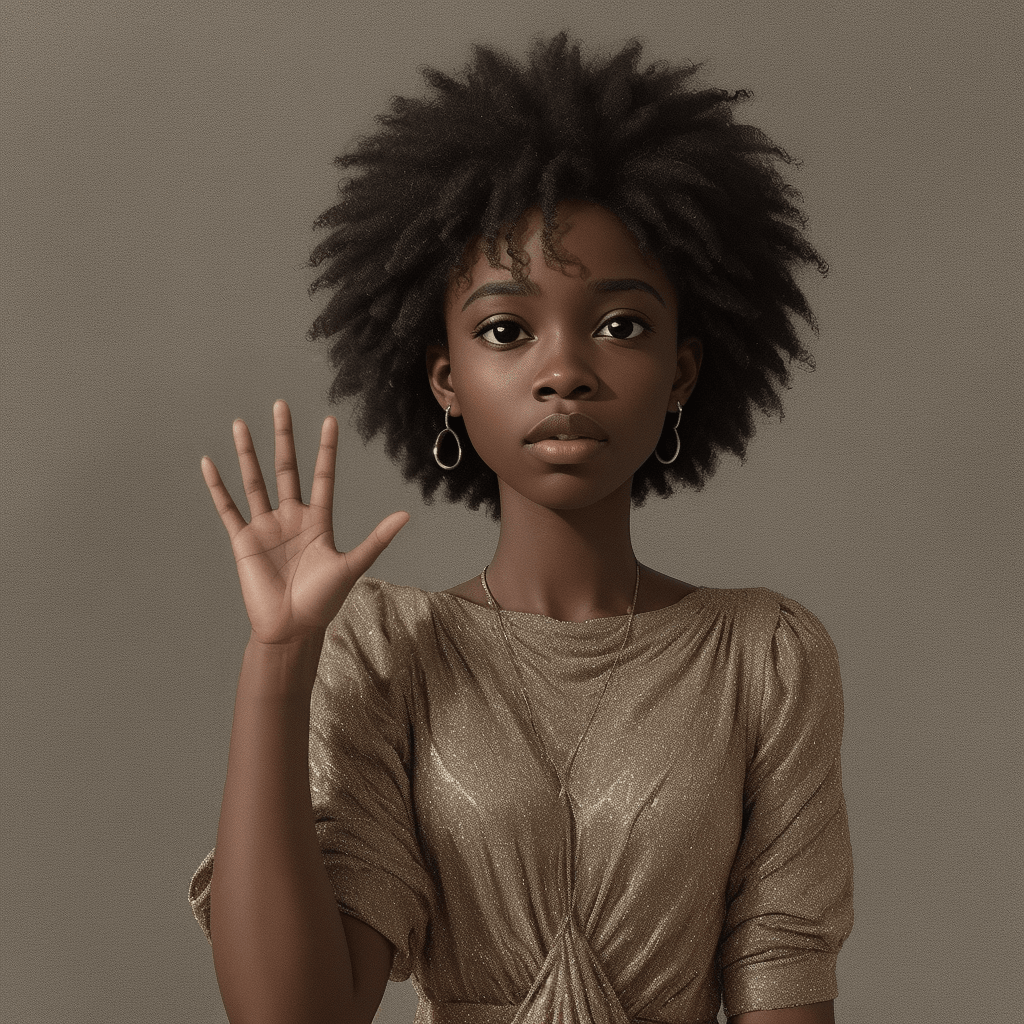}
  \caption*{\rev{PSNR: 59.81 dB / SSIM: 0.9996}}
\end{subfigure}%
% \vspace{-2pt}
\caption{Comparison between various hand restoration techniques. HandCraft demonstrates superior performance in hand rectification tasks, seamlessly integrating corrected hand into the original portraits. PSNR and SSIM metrics reveals that HandCraft achieves this while minimizing perturbations to non-hand regions, significantly outperforming the baseline HandRefiner~\cite{handrefiner_2023} method.}
\label{fig:hand_comparison}
\end{figure*}

\rev{
To demonstrate the effectiveness of HandCraft, we compare its performance with several ControlNet \cite{controlnet_2023} baselines, including ControlNet Hand Pose (CtrlNetHP), and ControlNet Segmentation (CtrlNetSeg), as well as the current state-of-the-art, HandRefiner \cite{handrefiner_2023}, on the MalHand-realistic dataset.
}
\rev{
The results in \cref{tab:sota_comparison} demonstrate HandCraft's superior performance. It achieves the highest hand pose confidence scores, 
indicating more natural hand postures. HandCraft also outperforms the null intervention and ControlNet variants in hand classifier confidence, while being comparable to HandRefiner, suggesting realistic, non-malformed hand restorations. Moreover, HandCraft attains the highest PSNR and SSIM scores in masked non-hand regions, showcasing its ability to maintain image integrity while correcting hand anatomy. 
}
\rev{
The results are further supported by qualitative comparisons conducted on the MalHand-artistic dataset, as shown in \cref{fig:hand_comparison}. These visual examples demonstrate that HandCraft not only corrects the malformed hands but also does so with a seamless integration into the original portrait, surpassing the performance of HandRefiner~\cite{handrefiner_2023}. 
More visual results can be found in the supplement. 
}

\rev{\subsection{Ablation Study}}

\rev{
\begin{table}[!t]
\centering
\caption{\rev{
Ablation study.
We ablate the masking strategy, removing the bounding box mask $M_d$ and the template mask $M_t$ independently, and show that both are important for high-quality restorations.
We also ablate the template alignment components: scale (S), translation (T), rotation (R), and handedness (H) alignment, showing that each are required for restoration.
}
}
\label{tab:mask_ablation}
\begin{tabularx}{\linewidth}{ccccccCC}
\toprule
$M_d$ & $M_t$ & S & T & R & H & $\bar{c}_{\text{pose}}$ $\uparrow$ & $\bar{c}_{\text{classifier}}$ $\uparrow$ \\
\midrule
\xmark & \cmark & \cmark & \cmark & \cmark & \cmark & 0.63 & 0.11 \\
\cmark & \xmark & \cmark & \cmark & \cmark & \cmark & 0.72 & 0.24 \\
\cmark & \cmark & \xmark & \cmark & \cmark & \cmark & 0.43 &  0.02 \\
\cmark & \cmark & \cmark & \xmark & \cmark & \cmark & 0.38 & 0.00 \\
\cmark & \cmark & \cmark & \cmark & \xmark & \cmark & 0.59 & 0.21 \\
\cmark & \cmark & \cmark & \cmark & \cmark & \xmark & 0.77 & 0.25 \\
\cmark & \cmark & \cmark & \cmark & \cmark & \cmark & \textbf{0.79} & \textbf{0.25} \\
\bottomrule
\end{tabularx}
\end{table}
}

\rev{
As shown in \cref{tab:mask_ablation}, we ablate several components of our method.
We demonstrate that removing any component of our mask input, including the bounding box mask $M_d$ and the template mask $M_t$, results in poorer restoration quality.
We also show that removing any of the template alignment components (scale, translation, rotation, and handedness alignment) also results in significantly poorer performance.
See the supplement for a qualitative ablation.
}

\section{Discussion and Conclusion}

Despite advances in generative image models, the issue of malformed hands has persisted across generations of such models, even in recently-released state-of-the-art systems like Stable Diffusion XL~\cite{sdxl_2023} and Sora~\cite{sora_2024}. We provide visualizations in the supplement highlighting instances of anatomically incorrect hands produced by these latest models.
As such, we expect HandCraft to remain useful for some time, since this issue appears to be quite persistent.
Even if this challenge is overcome by a new generative model, our method provides functionality to change hand gestures and poses for creative control, offering utility beyond just correcting malformed hands.

Our HandCraft framework addresses the challenge of correcting malformed hands in images generated by text-to-image diffusion models. Through the use of a parametric hand model to guide an image editor, we achieve seamless anatomical corrections that integrate with the original image's aesthetics. Our approach is effective and accessible, requiring no additional training. The Malhand datasets further provide resources for training and benchmarking. Comparisons with HandRefiner~\cite{handrefiner_2023} showed that HandCraft surpassed it in restoring hand anatomy while maintaining the integrity of the rest of the image.

%%%%%%%%% REFERENCES
\clearpage
{\small
\bibliographystyle{ieee_fullname}
\bibliography{main}
}

\clearpage

\end{document}